%% file: main.tex
\documentclass[10pt,twocolumn,letterpaper]{article}

\usepackage{iccv}              

\usepackage{stmaryrd}
\usepackage{multirow}   
\usepackage{pifont}
\usepackage{makecell}
\usepackage{graphics,dblfloatfix}
\usepackage[accsupp]{axessibility}

\definecolor{iccvblue}{rgb}{0.21,0.49,0.74}
\usepackage[pagebackref,breaklinks,colorlinks,allcolors=iccvblue]{hyperref}

\def\paperID{13954} 
\def\confName{ICCV}
\def\confYear{2025}

\title{Sequential keypoint density estimator: \\
an overlooked baseline of skeleton-based video anomaly detection}

\author{Anja Delić, Matej Grcić, Siniša Šegvić \\
University of Zagreb,
Faculty of Electrical Engineering and Computing \\
Zagreb, Croatia\\
{\tt\small name.surname@fer.hr}
}

\begin{document}
\maketitle
\input{sec/0_abstract}    
\input{sec/1_intro}

\input{sec/2_related_work}

\input{sec/3_method}
\input{sec/4_experiments}
\input{sec/5_conclusion}

\input{sec/6_acknowledgement}
{
    \small
    \bibliographystyle{ieeenat_fullname}
    \bibliography{main}
}

\input{appendix_content}
\end{document}

%% file: sec/0_abstract.tex
\begin{abstract}

Detecting anomalous human behaviour
is an important visual task
in safety-critical applications
such as healthcare monitoring,
workplace safety,
or public surveillance.
In these contexts,
abnormalities are often reflected
with unusual human poses.
Thus, we propose SeeKer,
a method for detecting anomalies
in sequences of human skeletons.
Our method formulates 
the skeleton sequence density
through autoregressive factorization at the keypoint level.
The corresponding conditional distributions
represent probable keypoint locations 
given prior skeletal motion.
We formulate the joint distribution of the considered skeleton
as causal prediction of conditional Gaussians
across its constituent keypoints.
A skeleton is flagged as anomalous 
if its keypoint locations surprise our model
(i.e.~receive a low density).
In practice, 
our anomaly score is a weighted sum 
of per-keypoint log-conditionals,
where the weights account 
for the confidence 
of the underlying keypoint detector.
Despite its conceptual simplicity,
SeeKer surpasses all previous methods
on the UBnormal and MSAD-HR datasets
while delivering competitive performance
on the ShanghaiTech dataset.

\end{abstract}

%% file: sec/1_intro.tex
\section{Introduction}
\label{sec:intro}

Anomaly detection in videos is a safety-critical computer vision task that focuses on identifying events that deviate from regular patterns within observed video sequences \cite{mo14icip,leyva17tip}.
This task is particularly relevant for applications such as public institution monitoring, workplace safety, or crowd surveillance \cite{sultani18cvpr,liu18cvpr}, where abnormal frames may indicate emergencies such as strokes, accidents or acts of violence \cite{acsintoae22cvpr}.
Designing more accurate human-related anomaly detectors may improve the security of shared spaces and contribute to overall public safety.

\begin{figure}[t]
    \centering
    \includegraphics[width=\linewidth]{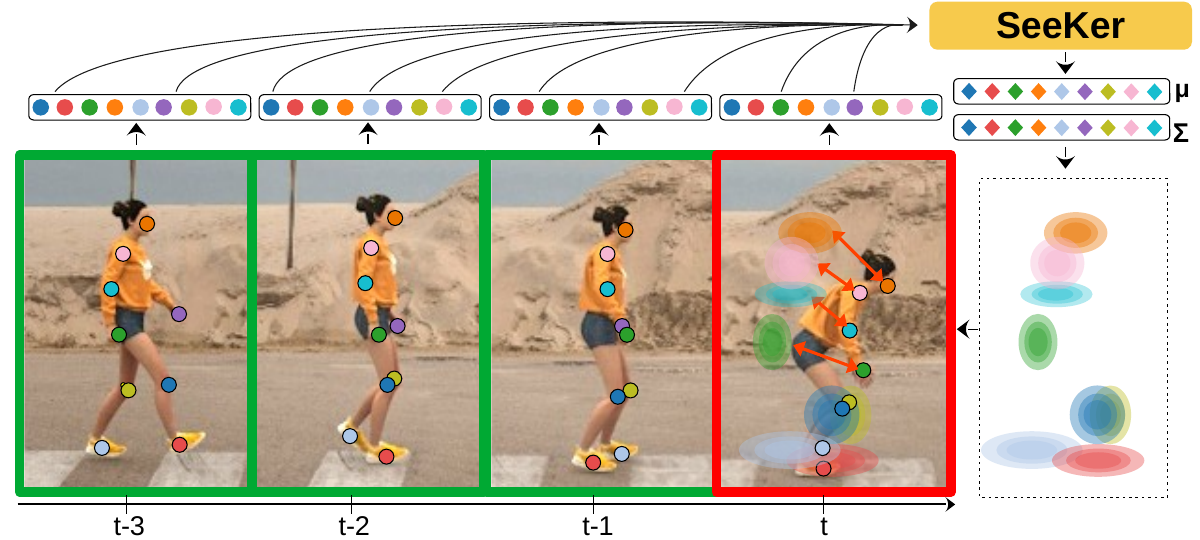}
    \caption{
    Given preceding skeleton keypoints (colored dots), SeeKer estimates the multivariate normal distribution of probable locations for the subsequent keypoint.
    Observing keypoints at improbable locations indicates anomalous behaviour (\textit{e.g.}~fainting).
    }
    \label{fig:motivation}
\end{figure}

Capturing normality in raw RGB video sequences is challenging due to high resolution, complex backgrounds, and spatio-temporal correlations between consecutive frames \cite{liu18cvpr}.
Consequently, previous anomaly detection approaches turn to feature representations extracted with deep models \cite{abati2019cvpr, yan23iccv} and optical flow predictions \cite{liu2021iccv} as well as to combinations of multiple modalities \cite{micorek24cvpr}. Although these approaches achieve varying degrees of success, video anomaly detection remains an open challenge.
In the context of human-related anomalies, promising approaches rely solely on skeletal representations of human poses \cite{flaborea2023iccv, hirschorn2023iccv}.
These are particularly relevant as problematic human behaviour can often be associated with irregular body positions \cite{acsintoae22cvpr}.
Moreover, skeleton sequences provide a low-dimensional and anonymous representation that reduces susceptibility to biases such as variations in clothing, body shape, or ethnicity and ensures that anomaly detection is based solely on motion dynamics.
Previous approaches to skeleton-based anomaly detection
\cite{luo2020neurcomp,yan23iccv,flaborea2024pr}
consider skeletons as graphs and train graph-oriented models \cite{yan18aaai} to maximize the likelihood of the skeleton sequence \cite{hirschorn2023iccv}.
However, such monolithic approaches ignore the causality in skeleton sequences inherent in the temporal domain and overlook the compositional nature of skeletons.
Moreover, anomaly scores of these approaches cannot account for the uncertainty of skeleton keypoint detectors.

In this manuscript, we propose SeeKer (\textbf{Se}qu\textbf{e}ntial \textbf{Ke}ypoint Density Estimato\textbf{r}),
a method for detecting anomalous human behaviour by predicting the conditional density of skeleton keypoints, as visualized in Figure~\ref{fig:motivation}.
Given the preceding keypoints, SeeKer predicts a multivariate normal distribution over possible locations for the subsequent keypoint.
Applying this approach across the entire skeleton sequence results in an autoregressive factorization of the joint sequence density at the keypoint level.
Consequently, SeeKer training corresponds to maximizing sequence likelihood.
During inference, SeeKer identifies skeletons as anomalous if their constituent keypoints deviate from the predicted distributions.
The corresponding skeleton anomaly score is a sum of keypoint likelihoods weighted by the keypoint extraction confidence from an off-the-shelf skeleton extractor.
Thus, our final decision accounts for uncertainties across all pipeline components.
Overall,
SeeKer presents the following key novelties
to skeleton-based anomaly detection:
autoregressive density factorization at the keypoint level, interpretable compositional anomaly scoring, and principled inclusion of keypoint detection uncertainty.
Experimental evaluation shows that
SeeKer achieves six percentage points absolute improvement over the best previous method on the UBnormal dataset and five percentage points absolute improvement on the MSAD-HR dataset in terms of AUROC.

%% file: sec/2_related_work.tex
\section{Related work}
\label{sec:related_work}

\noindent
\textbf{Video anomaly detection.}
Video anomaly detection \cite{mo14icip,leyva17tip} focuses on identifying anomalous frames within video sequences.
This task is particularly relevant in surveillance \cite{sultani18cvpr} where anomalies may indicate accidents or emergencies \cite{liu18cvpr}.
Early approaches to frame-based video anomaly detection map input frames into a shared feature space \cite{santos19vcir,abati2019cvpr} and apply decision rules that differentiate between normal and abnormal patterns \cite{ionescu2017iccv}.
More recent approaches \cite{ flaborea2023iccv,georgescu2022tpami,barbalau2022cviu} focus on bounding box detections and recognize anomalies using self-supervised features \cite{georgescu2020cvpr,wang2022eccv}.
Other methods leverage failed optical flow predictions \cite{liu2021iccv}, errors in denoising of diffusion models \cite{yan23iccv}, and predictions from cooperatively trained deep models \cite{zaheer2022cvpr}. 
In addition, there are exciting density estimation techniques such as denoising score matching \cite{micorek24cvpr}.
However, these RGB-based methods may raise ethical concerns due to potential appearance-based biases and difficult anonymization.
SeeKer builds atop skeleton sequences, and thus avoids such 
concerns.

\noindent
\textbf{Skeleton-based video anomaly detection.}
Video anomaly detection based on human skeleton sequences \cite{markovitz2020cvpr, morais2019cvpr} is particularly effective due to anonymity, low computational requirements, and domain invariance \cite{hirschorn2023iccv}.
Existing skeleton-based methods capture spatio-temporal dependencies within skeleton sequences using
recurrent models \cite{morais2019cvpr} or
convolutional variational autoencoders \cite{miracle2022arxiv, jain2020icpr}.
Alternatively, viewing skeletons as graphs \cite{markovitz2020cvpr,luo2020neurcomp, flaborea2024pr,hirschorn2023iccv} allows for processing with graph-based architectures \cite{yan18aaai, yu18ijcai}.
The trained models then detect anomalies as failed pose predictions \cite{jain2020icpr,flaborea2023iccv} or unlikely sequences \cite{hirschorn2023iccv}.
Unlike previous approaches, SeeKer
leverages the compositional nature of skeletons and applies autoregressive sequence factorization at the keypoint level.
This results in improved alignment,
principled inclusion of keypoint detection uncertainty,
and interpretable scoring in the sense that
we can easily identify the keypoint(s) 
that triggered the anomaly detector.




\noindent
\textbf{Human pose estimation.}
Human pose estimation \cite{sapp10eccv,sun12cvpr} aims to detect key body joints in an image and assemble them into skeletal structures. 
Recent pose estimators \cite{fang23tpami, zhu24tpami} allow for accurate real-time detection of all skeletons in a given image, facilitating skeleton-based recognition.
SeeKer builds on efficient pose estimators \cite{fang23tpami} to enable real-time anomaly detection.
An overview of human pose estimation can be found in surveys \cite{chen20cviu, lan23hms}.

\noindent
\textbf{Autoregressive data likelihood factorization.}
Autoregressive factorization of data likelihood emerged as a foundational approach across numerous machine learning applications with a dominant temporal domain \cite{berezina10icassp,devlin19naacl}.
Such factorization is conveniently modeled using sliding window based approaches \cite{waibel89assp,braverman15book}, recurrent neural networks \cite{rumelhart86n,hochreiter97nc} or
transformer decoders \cite{vaswani17nips,carion20eccv} with causal self-attention layers \cite{tarzanaghL23neurips,li24aistats}.
SeeKer relies on autoregressive factorization of skeleton sequence density.

%% file: sec/3_method.tex
\section{Skeleton anomaly detection by predicting probable locations of subsequent keypoint}
\label{sec:method}
\textbf{Problem setup.}
Let $\mathcal{D} = \{\mathbf{X}^i 
 = (X_1^i, X_2^i, \dots,  X_{T_i}^i)\}_{i=1}^M$ be a dataset of $M$ skeleton sequences $\mathbf{X}^i$ that track human activity throughout $T_i$ time steps.
Each skeleton $X^i_t \in \mathbf{X}^i$ is extracted from the corresponding image and tracked across multiple frames with AlphaPose \cite{fang23tpami}, an off-the-shelf detector and tracker.
Each skeleton $X_t$ is an $N\times D$ matrix with rows corresponding to skeleton joints and columns representing the spatial coordinates of the detected keypoint.
In practice, $N=18$ (number of keypoints) and $D = 2$ (coordinates of each keypoint).
Thus, each sequence $\mathbf{X}^i$ forms a third-order tensor of shape $T_i  \times N \times D$.
Given the dataset $\mathcal{D}$ of normal skeleton sequences, our objective is to develop an anomaly detector $\tau: \mathbf{X}^i \rightarrow \{0, 1\}^{T_i}$ capable of distinguishing between regular and anomalous skeletons in each time instant of a given sequence.

\begin{figure*}[htb]
    \centering
    \includegraphics[width=\linewidth]{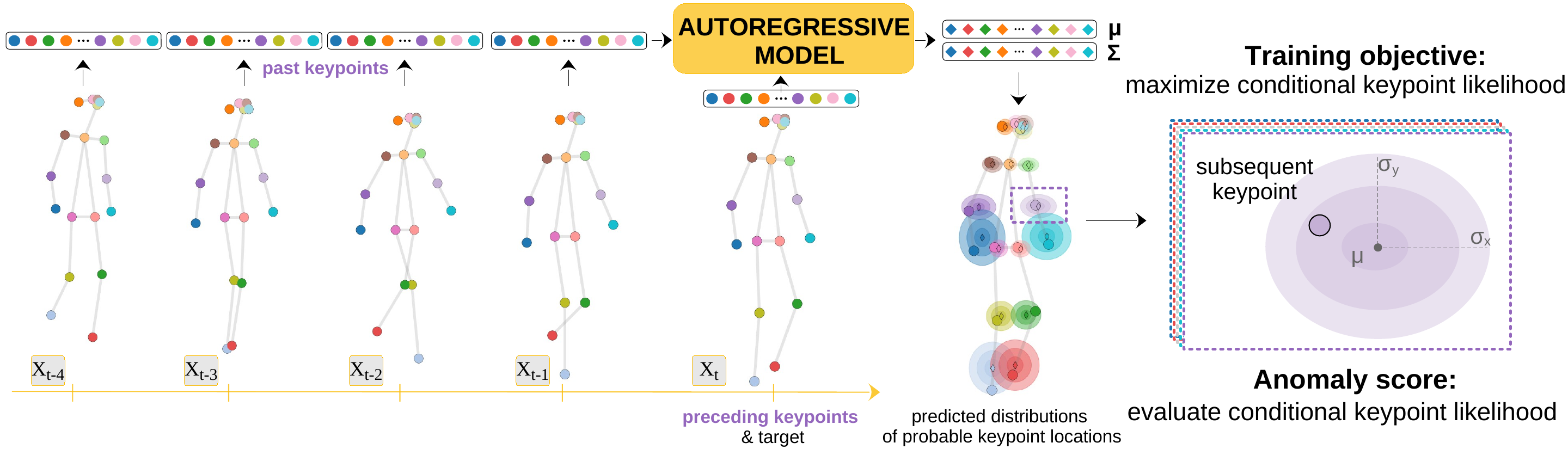}
    \caption{SeeKer training starts by flattening the target skeleton $X_t$ and the preceding skeletons $X_{<t}$ into vector representations. 
    Then, the concatenated vectors of the preceeding keypoints and the target skeleton are fed to an autoregressive model, which outputs the mean $\boldsymbol\mu_\theta$ and covariance $\Sigma_\theta$ for each keypoint of the target.
    These values induce normal distributions under which each keypoint should be likely.
    }
    \label{fig:method_figure}
\end{figure*}
The core approach behind SeeKer
is to predict a conditional distribution of probable keypoint locations based on the preceding keypoints in the skeleton sequence, as visualized in Figure~\ref{fig:method_figure}.
Leveraging this design in conjunction with the compositional nature of skeletons
provides a highly interpretable skeleton anomaly score that aggregates likelihoods estimated at the keypoint level.

\subsection{Autoregressive density of skeleton sequences}

We derive the conditional per-keypoint density from the joint density of a skeleton sequence with a top-down approach.
The joint density of a skeleton sequence $\mathbf{X}$
can be conveniently factorized
by autoregressive modeling \cite{germain15icml, brown20neurips} in order to capture skeleton motion:
\begin{equation}
\label{eq:likelihood}
    p_\theta(\mathbf{X}) = \prod_{t=1}^T  p_\theta(X_t| \mathbf{X}_{<t}) \approx \prod_{t=1}^T  p_\theta(X_t| \mathbf{X}_{\Delta}).
\end{equation}
Here, the conditional likelihood of a skeleton $X_t$ 
is determined from the skeletons from the past $t-1$ time steps, denoted by $\mathbf{X}_{<t}$.
Given skeleton sequences of arbitrary length, we adopt a siding-window approach with a fixed-size time window. 
In practice, this means that we can faithfully approximate $p_\theta(X_t | \mathbf{X}_{<t})$ using only past $\Delta$ skeletons, i.e., $p_\theta(X_t | \mathbf{X}_{\Delta})$, 
where $\Delta$ is sufficiently large to capture crucial temporal dependencies.

\begin{figure}[ht]
    \centering
    \includegraphics[width=\linewidth]{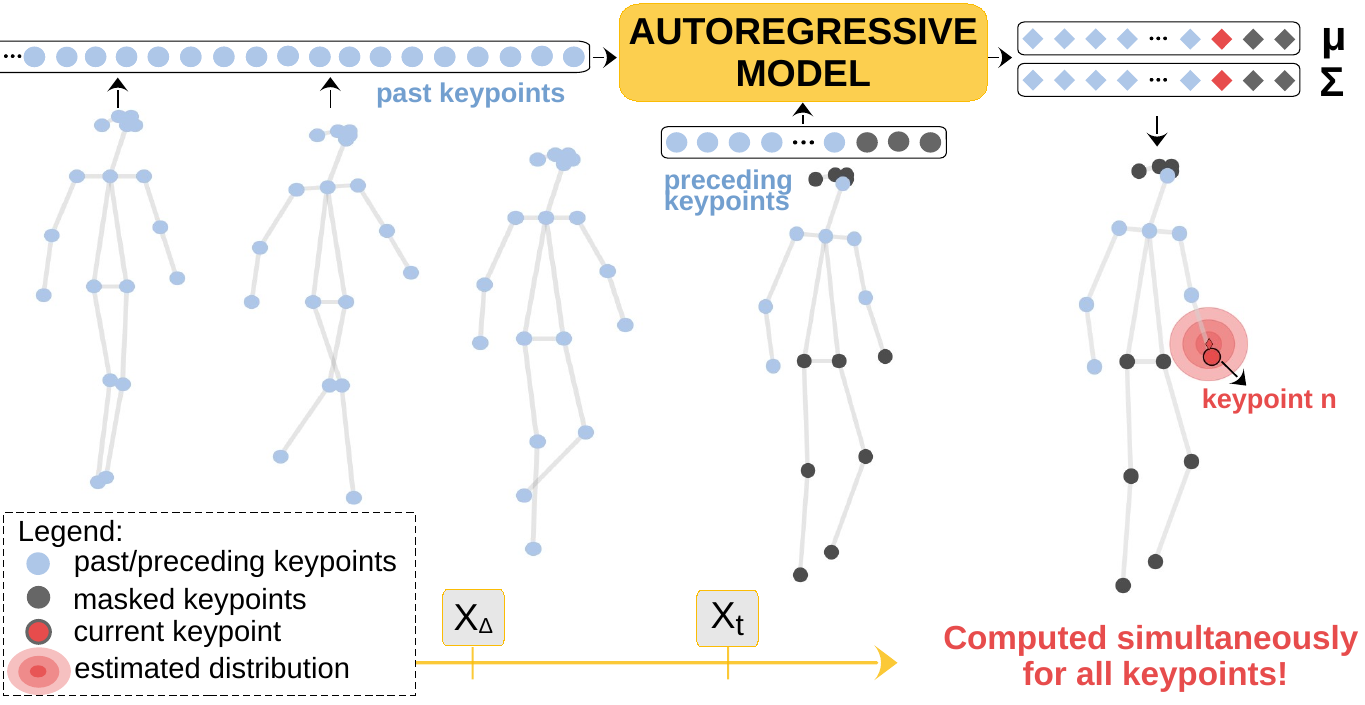}
    \caption{
    We autoregressively predict distributions of possible keypoint locations for the current skeleton $X_{t}$.
    For instance, the distribution of the current keypoint $X_{t,n}$ (denoted in red) is estimated from all past keypoints $X_\Delta$ and the preceding keypoints $X_{t, <n}$, all denoted in blue. Dark keypoints are masked due to autoregressive likelihood factorization.}
    \label{fig:masking}
\end{figure}

Since each skeleton consists of keypoints, we can further model the conditional likelihood of each skeleton $X_t$ as the autoregressive joint likelihood of $N$ keypoints:
\begin{equation}
\label{eq:skel_joi_lik}
    p_\theta(X_t|\mathbf{X}_{\Delta}) = \prod_{n=1}^N p_\theta(X_{t,n}|X_{t,<n},\mathbf{X}_{\Delta}).
\end{equation}
We condition the density of $n$-th keypoint at timestep $t$ on the preceding $n-1$ keypoints of the current timestep, as well as  $N \cdot \Delta$ keypoints from the past timesteps, as ilustrated on Figure~\ref{fig:masking}.
This formulation requires a fixed ordering of keypoints across skeletons. However, as experimentally confirmed later, keypoint ordering does not affect the model's expressiveness \cite{malach24icml}.


We further define per-keypoint conditional likelihoods as multivariate normal distributions with mean $\boldsymbol\mu_\theta$ and covariance  $\Sigma_\theta$ computed from the previous keypoints:
\begin{equation}
\label{eq:normal_fact}
    p_\theta(X_{t,n}|X_{t,<n},\mathbf{X}_{\Delta}) := \mathcal{N}(X_{t,n}|\boldsymbol{\mu}_\theta, \Sigma_\theta).
\end{equation}
Both
$\boldsymbol{\mu}_\theta$ and $\Sigma_\theta$ are predicted from the preceding keypoints in the current timestep $X_{t,<n}$ and the past keypoints $\mathbf{X}_{\Delta}$ using a deep autoregressive model with tuneable parameters $\theta$.
Altogether, we can now express the log-likelihood of the entire sequence (\ref{eq:likelihood}) in terms of the keypoint likelihoods:
\begin{equation}
\label{eq:seq_keypoint_loglik}
    \ln p_\theta(\mathbf{X}) = \sum_{t=1}^T  \sum_{n=1}^N \ln \mathcal{N}(X_{t,n}|\boldsymbol{\mu}_\theta, \Sigma_\theta).
\end{equation}

\noindent
\textbf{Optimization objective.}
Given a dataset of regular skeleton sequences  $\mathcal{D}$, we recover the maximum likelihood estimate of the parameters $\theta$ by minimizing:
\begin{equation}
   \theta_\text{MLE} =  \underset{\theta}{\text{argmax}} \,L(\theta; \mathcal{D}),  L(\theta; \mathcal{D}) = -\sum_{\mathbf{X} \in \mathcal{D}} \ln p_\theta(\mathbf{X}) .
\end{equation}
Due to Gaussian nature of our distributions and autoregressive factorization of the sequence likelihood (\ref{eq:seq_keypoint_loglik}), the objective $L(\theta; \mathcal{D})$ simplifies to (as detailed in Appendix \ref{app:derivation}):
\begin{equation}
\label{eq:loss_objective}
    L(\theta; \mathcal{D})
    \cong \sum_{\mathbf{X}, t, n} (X_{t, n} - \boldsymbol{\mu}_\theta)^\top\Sigma^{-1}_\theta(X_{t, n} - \boldsymbol{\mu}_\theta) + \ln \text{det} \, \Sigma_\theta .
\end{equation}
The objective (\ref{eq:loss_objective}) encourages the model to minimize the Mahalanobis distance \cite{md18ssa} between the observed keypoint $X_{t, n}$  and the normal distribution with predicted parameters $\boldsymbol\mu_\theta$ and $\Sigma_\theta$.
The second loss term can be viewed as a form of regularization that prevents trivial solutions, \textit{e.g.}~large covariance that makes all locations equally likely.



Figure~\ref{fig:method_figure} visualizes the training procedure of SeeKer.
We aggregate locations from the preceding keypoints
and feed them to an autoregressive model that estimates the distribution parameters ($\boldsymbol\mu$ and $\Sigma$) for each of the subsequent keypoints in the target skeleton.
The training proceeds by adapting the predicted distribution parameters so that the likelihood of the observed keypoints is maximized.
Note that the presented design offers great interpretability since it operates in the original 2D space of skeleton keypoints.

\begin{figure}[ht]
    \centering
    \includegraphics[width=\linewidth]{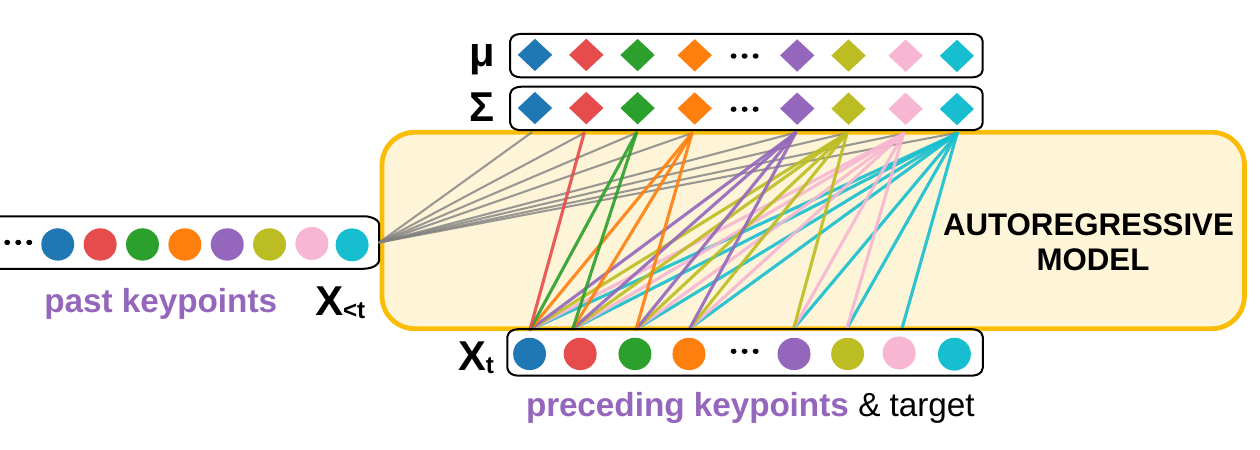}
    \caption{Graphical representation of the causality constraints in autoregressive factorization. The first (blue) output depends only on past keypoints, while the green output depends on all past keypoints and preceding blue and red keypoints from the target.
    }
    \label{fig:autoreg}
\end{figure}

\noindent
\textbf{Autoregressive architectures.}
The nature of skeleton sequences implies that we will anticipate $N$ new keypoints at every timestep $t$.
Therefore, an appropriate model must be capable to simultaneously predict the mean $\boldsymbol\mu$ and covariance $\Sigma$ for $N$ multivariate normal distributions.
Figure \ref{fig:autoreg} illustrates the required relationships between inputs and outputs in the SeeKer framework.
Observe that these relations closely resemble the teacher forcing computational graph \cite{lamb16neurips} commonly used in natural language processing.

The described causal constraint can be enforced using different architectures \cite{hochreiter97nc,germain15icml, vaswani17nips}. 
Our experimental evaluation revealed that the relatively simple masked fully connected model \cite{germain15icml} performs best for our task,
which can be attributed to the uniformity of the skeleton sequences.
Our model differs from \cite{germain15icml} by following the natural ordering of inputs while allowing neighboring inputs (e.g., keypoint coordinates) to be codependent rather than strictly causal. Thus, we enforce causality through fixed blockwise triangular masks.
We revisit this architecture in Appendix \ref{app:architectures}.


%

\subsection{Composite anomaly score}
\label{sec:method-score}
SeeKer expresses the density of a skeleton sequence as aggregate key-point density (Eq.~\ref{eq:skel_joi_lik}).
Thus, observing a keypoint at a probable location indicates normal (expected) behavior.
Contrary, observing the keypoint at an improbable location signals abnormal behavior.
We define the anomaly score of the skeleton $X_t$ given the context $\mathbf{X}_{\Delta}$ as a joint density of its keypoints:
\begin{equation}
    s'(X_t|\mathbf{X}_{\Delta}) = - \sum_{n} \ln p_{\theta_\text{MLE}}(X_{t,n}|X_{t,<n},\mathbf{X}_{\Delta}).
\label{eq:skeleton_score_inial}
\end{equation}
The anomaly score (\ref{eq:skeleton_score_inial}) assumes that the entire skeleton $X_t$ is accurately extracted from the corresponding scene $\mathcal{I}_t$.
However, occlusions and faulty skeleton detectors may produce errors.
Fortunately, most detectors provide a confidence score for each keypoint \cite{fang23tpami}.
Consequently, we introduce 
the detection confidence
into our anomaly score as:
\begin{equation}
    s(X_t|\mathbf{X}_{\Delta}) = - \sum_{n} 
 c_{t, n}  \ln p_{\theta_\text{MLE}}(X_{t,n}|X_{t,<n},\mathbf{X}_{\Delta}).
\label{eq:anomaly_score}
\end{equation}
The score (\ref{eq:anomaly_score}) weighs each keypoint log-density by its detection confidence $c_{t,n} \in [0,1]$. 
Thus, our anomaly score integrates the
uncertainties of all pipeline components.
Given a threshold $\delta$, we can classify each frame as $\tau(X_t) = \llbracket s(X_t|\mathbf{X}_{<t}) > \delta \rrbracket$, where  $\llbracket \cdot \rrbracket$ represents Iverson brackets.
Note that the proposed anomaly score
readily discloses per-keypoint 
contributions to the decision,
which makes it much more interpretable than the current state of the art \cite{hirschorn2023iccv, micorek24cvpr}.


In the case when multiple humans are present in a scene $\mathcal{I}_t$, we simply determine the per-frame anomaly score as the maximum score among all skeletons within that frame.
Consequently, our per-frame anomaly score $ s_F$ equals:
\begin{equation}
    s_F(\mathcal{I}_t |\mathcal{I}_{<t} ) = \max_{X_t \in  \mathcal{I}_t} s(X_t|\mathbf{X}_{\Delta}).
\end{equation}
Next, we experimentally validate SeeKer utility on the standard human-related anomaly detection benchmarks.

%% file: sec/4_experiments.tex
\section{Experimental setup}
\label{sec:experimental_setup}


\noindent
\textbf{Datasets.}~We consider three publicly available datasets for unsupervised video anomaly detection in skeleton sequences.
UBnormal \cite{acsintoae22cvpr} is a synthetic dataset with 268 training, 64 validation, and 211 test video sequences.
The inlier sequences contain actions such as walking while texting and talking to others.
Anomalous actions include unexpected behaviour, such as stealing or fighting.
This is the most relevant dataset for our task due to highly accurate labels.
ShanghaiTech \cite{liu18cvpr} consists of 330 training and 107 test videos of real-world scenes. 
Walking, standing, or sitting are considered as inlier actions, while cycling, skateboarding, and fighting are anomalous. Furthermore, both datasets contain anomalous events that are not human-related (e.g.~car accident or fire).
Consequently, we also consider dataset subsets that contain only human-related anomalies as suggested in \cite{flaborea2023iccv,rodrigues2020wacv} refered to as UBnormal-HR and ShanghaiTech-HR.
The MSAD \cite{zhu25neurips} dataset includes videos captured across diverse locations and various viewpoints, ensuring a wide range of perspectives and conditions, making it the most challenging benchmark to date. 
MSAD contains 720 videos in total, some extending up to 6K frames, and features 55 distinct abnormal events.
We focus on the human-related subset with 20 distinct abnormal actions, MSAD-HR, as the full dataset track includes scenarios and events related to industrial automation and manufacturing without any human involvement.
Note that non-human related anomalies on UBnormal and ShanghaiTech still involve humans, making them detectable through human reactions to events like fire or unusual vehicle behavior.
Previous work \cite{hirschorn2023iccv} noticed that contemporary skeleton extractors \cite{fang23tpami} and trackers \cite{xiu18bmvc} fail to produce satisfactory outputs on some datasets (UCFCrime \cite{sultani18cvpr} and Avenue \cite{lu13iccv}).
Hence, we skip these datasets in our experiments.

\noindent
\textbf{Metrics.}~We report the standard anomaly detection metrics AUROC and AP, as well as human-related anomaly detection metrics RBDC and TBDC. Notably, we are the first to introduce the measurement of AP in this context.
We compute AUROC and AP jointly over all test frames, which therefore corresponds to micro-averaging, \textit{e.g.}~micro-AUROC.  
We also report the region-based and track-based detection criteria (RBDC and TBDC) \cite{ramachandra20wacv} to evaluate the spatial and temporal localization of detected anomalies on UBnormal and ShanghaiTech since they offer location annotations.
RBDC assesses each detected region by calculating its Intersection-over-Union (IoU) with the corresponding ground truth region, labeling it as a true positive if the overlap exceeds a specified threshold $\alpha$. 
On the other hand, TBDC evaluates the accuracy of tracking abnormal regions over time, identifying a detected track as a true positive if the number of detections within the track exceeds a threshold $\beta$. 
Following the previous work \cite{barbalau2022cviu}, we set both $\alpha$ and $\beta$ at 0.1.
We calculate the area under the ROC curve for both metrics, taking into account the trade-off between detection rate and false-positive rate.
For this evaluation, we rely on the ground truths provided by \cite{georgescu2022tpami}.
Some methods are not suitable for calculating RBDC and TBDC since they cannot pair features to their temporal or spatial locations in the original videos \cite{micorek24cvpr}.

\noindent
\textbf{Baselines.}~We organize the considered baselines
with respect to the modality they receive on the input.
Methods that operate on deep
features \cite{liu18cvpr,hasan2016cvpr} are denoted with (D).
Methods that utilize optical flow velocities \cite{liu2021iccv,yu18ijcai} are denoted with (V), while skeletons-based approaches \cite{hirschorn2023iccv,micorek24cvpr} are denoted with (S).
We briefly discuss the most relevant skeleton-based methods.
STG-NF \cite{hirschorn2023iccv} proposes to estimate the likelihood of skeleton sequences with a graph-oriented normalizing flow with space and time separable graph convolutions in the coupling layers. 
MoCoDAD \cite{flaborea2023iccv} proposes to model both normality and abnormality as multimodal by forecasting future skeletons with a diffusion model conditioned on the past.
MULDE \cite{micorek24cvpr} is a recent method that leverages
an energy-based model trained by a modification of denoising score matching that requires the model to have two times differentiable layers.
For a fair comparison, we evaluate the AP of the most relevant baseline, STG-NF, with models provided by the authors.

\noindent
\textbf{Implementation details.}
Our skeleton extractor procedure closely follows STG-NF \cite{hirschorn2023iccv}.
We extract human skeletons with AlphaPose \cite{fang23tpami} within the YOLOX bounding boxes \cite{ge21arxiv} and track skeletons with reID \cite{zhou19iccv} across the video. The output skeletons consist of 18 keypoints in the COCO pose format \cite{lin14eccv}.
We divide the extracted skeleton trajectories into time segments of $T=24$ frames and normalize them to have zero mean and unit variance.

We predict the per-keypoint distribution parameters $\boldsymbol\mu$ and $\boldsymbol\Sigma$ with deep autoregressive models that we configure for skeleton sequences.
We consider causally masked fully connected models \cite{germain15icml}, transformers \cite{vaswani17nips}, and recurrent models.
We train the causal fully connected model for 20 epochs on UBnormal and MSAD-HR  and 10 epochs on ShanghaiTech with the Adam optimizer, learning rate $10^{-3}$ and batch size 256.
Since UBnormal is the only dataset with a validation split, we validate all hyperparameters on the UBnormal validation subset and reuse them for ShanghaiTech and MSAD-HR.
During inference, SeeKer considers $T$ preceding frames in a sliding-window manner.
At the beginning of a sequence, we backtrack the score of the $T$-th frame to the first $T-1$ frames.
Following \cite{hirschorn2023iccv, flaborea2023iccv, micorek24cvpr}, we finally apply Gaussian smoothing of the computed anomaly scores.
However, we apply smoothing for each person separately before aggregating the per-frame anomaly score with $\sigma$ = 10.
All reported results are averaged over three runs, and variance is consistently below $0.05$ pp. 
Our code is
publicly available\footnote{\url{https://github.com/adelic99/seeker}}.

\section{Experimental results}
\label{sec:experiments}

\subsection{Skeleton-based video anomaly detection}

Table~\ref{tab:ubnormal} compares SeeKer with previous methods on the UBnormal dataset.
Our method substantially improves over all previous approaches in terms of AUROC, AP, and RBDC.
Compared to the best previous skeleton-based method STG-NF \cite{hirschorn2023iccv}, SeeKer attains substantial increase of over 6 percentage points (pp) AUROC, 17 pp improvement in terms of AP,
and 10 pp improvement in terms of RBDC, all while maintaining competitive TBDC.
Moreover, SeeKer achieves over 15 pp AUROC improvement compared to the leading method based on deep features Ssmtl++ \cite{georgescu2020cvpr}.
SeeKer also outperforms the multimodal baseline MULDE \cite{micorek24cvpr}, despite relying solely on skeletons.
For human-related anomalies, SeeKer achieves over 7 pp gain in terms of AUROC and over 12 pp AP improvement over the best performing baseline STG-NF \cite{hirschorn2023iccv}.
Thus, SeeKer is a new best model for anomaly detection on the UBnormal dataset.
\begin{table}[ht]
\setlength{\tabcolsep}{2pt}
\footnotesize
\centering
  \begin{tabular}{ccc|lcccccc}
    \multicolumn{3}{c|}{Modality} & \multirow{2}{*}{Method} & \multicolumn{2}{c}{AUROC} & \multicolumn{2}{c}{AP} & \multirow{2}{*}{RBDC} & \multirow{2}{*}{TBDC}\\
    D & V & S &  & Full & HR & Full & HR &  &  \\
    \hline \rule{0pt}{1.2em}
    \checkmark & & & BAF \cite{georgescu2022tpami} & 59.3 & - & - & - & 21.9 & 53.4 \\
    \checkmark & & & Jigsaw \cite{wang2022eccv} & 56.4 & - & - & - & 11.8 & 36.8 \\
    \checkmark & & & Ssmtl++ 
    \cite{georgescu2020cvpr} & 62.1 & - & - & - & 25.6 & \textbf{63.5} \\
    \checkmark & \checkmark & & FPDM\cite{yan23iccv} & 62.7 & -  & - & - & - & - \\
    \checkmark & \checkmark & \checkmark & MULDE \cite{micorek24cvpr} & 72.8 & - & - & - & - & - \\[0.5em]
    & & \checkmark & BiPOCO \cite{miracle2022arxiv} & 50.7 & 52.3 & - & - & - & - \\
    & & \checkmark & GEPC \cite{markovitz2020cvpr} & 53.4 & 55.2 & - & - & - & - \\
    & & \checkmark & MPED-RNN \cite{morais2019cvpr} & 60.6 & 61.2 & - & - & - & - \\
    & & \checkmark & COSKAD \cite{flaborea2024pr} & 65.0 & 65.5 & - & - & - & - \\
    & & \checkmark & MoCoDAD \cite{flaborea2023iccv} & 68.3 & 68.4 & - & - & - & - \\
    & & \checkmark & STG-NF \cite{hirschorn2023iccv} & 71.8 & 71.5 & 62.7 & 67.2 & 31.7 & \textbf{62.3} \\
    \hline \rule{0pt}{1.2em}
    & & \checkmark & SeeKer & \textbf{77.9} & \textbf{78.9}  & \textbf{80.3} & \textbf{79.8} & \textbf{42.0} & \underline{58.5} \\
\end{tabular}
    \caption{Experimental evaluation of unsupervised anomaly detection on UBnormal \cite{acsintoae22cvpr}. Input modality acronyms denote deep features (D), optical flow (V), and skeletons (S).}
    \label{tab:ubnormal}
\end{table}

Table \ref{tab:shanghai_tech} compares SeeKer performance with all previous approaches on the ShanghaiTech dataset. 
Among skeleton-based methods, our approach attains the highest AP and RBDC, while ranking second in AUROC and TBDC.
Notably, the improvement upon the top-performing skeleton baseline STG-NF is over 9pp in terms of RBDC.
In the case of human-related anomalies, SeeKer achieves the best AP and the second-best AUROC among skeleton-based approaches and continues to outperform methods that depend on deep features and optical flow velocities.
Overall, SeeKer is a competitive model on ShanghaiTech dataset.
\begin{table}[ht]
\setlength{\tabcolsep}{2pt}
\footnotesize
\centering
    \begin{tabular}{ccc|lcccccc}
    \multicolumn{3}{c|}{Modality} & \multirow{2}{*}{Method} & \multicolumn{2}{c}{AUROC} & \multicolumn{2}{c}{AP} & \multirow{2}{*}{RBDC} & \multirow{2}{*}{TBDC}\\
    D & V & S &  & Full & HR & Full & HR &  &  \\
    \hline \rule{0pt}{1.2em}
    \checkmark & & & sRNN \cite{luo2017iccv} & 68.0 & - & - & - & - & - \\
    \checkmark & & & Conv-AE \cite{hasan2016cvpr} & 70.4 & 69.8 & - & - & - & - \\
    \checkmark & & & LSA \cite{abati2019cvpr} & 72.5 & - & - & - & - & - \\
    \checkmark & \checkmark & & FFP \cite{liu2018cvpr} & 72.8 & 72.7 & - & - & - & - \\
    \checkmark & & & GCL \cite{zaheer2022cvpr} & 79.6 & - & - & - & - & - \\
    \checkmark & \checkmark & & CAE-SVM \cite{ionescu2017iccv} & 78.7 & - & - & -  & - & -\\
    \checkmark & \checkmark & & VEC \cite{yu20acm} & 74.8 & - & - & - & - & - \\
    \checkmark & \checkmark & & $\text{HF}^2$ \cite{liu2021iccv} & 76.2 & - & - & - & - & - \\
    \checkmark & \checkmark & & FPDM\cite{yan23iccv} & 78.6 & - & - & - & - & - \\
    \checkmark & & & SSMTL \cite{georgescu2020cvpr} & 82.7 & - & - & - & - & - \\
    \checkmark & \checkmark & & BA-AED \cite{georgescu2022tpami} & 82.7 & - & - & - & 41.3 & 78.8 \\
    \checkmark & & & SSMTL++ \cite{barbalau2022cviu} & 83.8 & - & - & - & 47.1 & 85.6 \\
    \checkmark & & & Jigsaw \cite{wang2022eccv} & 84.2 & 84.7 & - & - & 22.4 & 60.8 \\
    \checkmark & \checkmark & \checkmark & MULDE \cite{micorek24cvpr} & \textbf{86.7} & - & - & - & 52.7 & \textbf{83.6} \\[0.5em]
    & & \checkmark & BiPOCO \cite{miracle2022arxiv} & - & 74.9 & - & - & - & - \\
    & & \checkmark & MPED-RNN \cite{morais2019cvpr} & 73.4 & 75.4 & - & - & - & - \\
    & & \checkmark & MTP \cite{rodrigues2020wacv} & 76.0 & 77.0 & - & - & - & - \\
    & & \checkmark & GEPC \cite{markovitz2020cvpr} & 76.1 & 74.8 & - & - & - & - \\
    & & \checkmark & PoseCVAE \cite{jain2020icpr} & - & 75.5 & - & - & - & - \\
    & & \checkmark & Normal Graph \cite{luo2020neurcomp} & - & 76.5 & - & - & - & - \\
    & & \checkmark & COSKAD \cite{flaborea2024pr} & - & 77.1 & - & - & - & - \\
    & & \checkmark & GCAE-LSTM \cite{li2021neurocmp} & - & 77.2 & - & - & - & - \\
    & & \checkmark & MoCoDAD \cite{flaborea2023iccv} & - & 77.6 & - & - & - & - \\
    & & \checkmark & MULDE \cite{micorek24cvpr} & 78.5 & - & - & - & - & - \\
    & & \checkmark & STG-NF \cite{hirschorn2023iccv} & \textbf{85.9} & \textbf{87.4}  & 77.6 & 81.4 & 52.1 & \textbf{82.4} \\
    \hline \rule{0pt}{1.2em}
    & & \checkmark & SeeKer & \underline{85.5} & \underline{86.9} & \textbf{80.0} & \textbf{81.5} & \textbf{62.5} & \underline{81.8} \\
    \end{tabular}

    \caption{Experimental evaluation of anomaly detection on ShanghaiTech \cite{luo2017iccv}. 
    Input modality acronyms denote deep features (D), optical flow (V), and skeletons (S). 
    }
    \label{tab:shanghai_tech}
\end{table}

Table \ref{tab:mulde} compares SeeKer with relevant skeleton-based methods on the MSAD-HR dataset \cite{zhu25neurips}. 
As the first to report results on MSAD-HR, we reproduce the most relevant baselines \cite{flaborea2023iccv, hirschorn2023iccv}.
SeeKer surpasses these baselines by a substantial margin of 5.4 pp in AUROC and more than 3 pp in terms of AP.
The MSAD-HR dataset presents significant challenges due to crowded scenes, where frequent human trajectory intersections impact skeleton estimation and tracking accuracy.
Consequently, the baseline methods only marginally exceed random guessing.
Next, we qualitatively analyse SeeKer performance to gain deeper insights into the obtained quantitative results.

\begin{table}[ht]
\setlength{\tabcolsep}{4pt}
\footnotesize
\centering
  \begin{tabular}{ccc|lcc}
    \multicolumn{3}{c|}{Modality} & \multirow{2}{*}{Method} & \multirow{2}{*}{AUROC} & \multirow{2}{*}{AP}\\ 
    D & V & S &&&\\
    \hline
    && \checkmark &MoCoDaD \cite{flaborea2023iccv} & 53.9 & 51.7\\
    && \checkmark &STG-NF \cite{hirschorn2023iccv} & 55.7 & 56.5\\
    \hline 
    && \checkmark &SeeKer & \textbf{61.1 } & \textbf{ 60.1}
\end{tabular}
    \caption{Experimental evaluation of unsupervised anomaly detection on the MSAD-HR \cite{zhu25neurips}.
    All baseline evaluations were reproduced by us.
    }
    \label{tab:mulde}
\end{table}

\subsection{Qualitative performance}
\begin{figure}[htb]
    \centering
    \includegraphics[width=\linewidth]{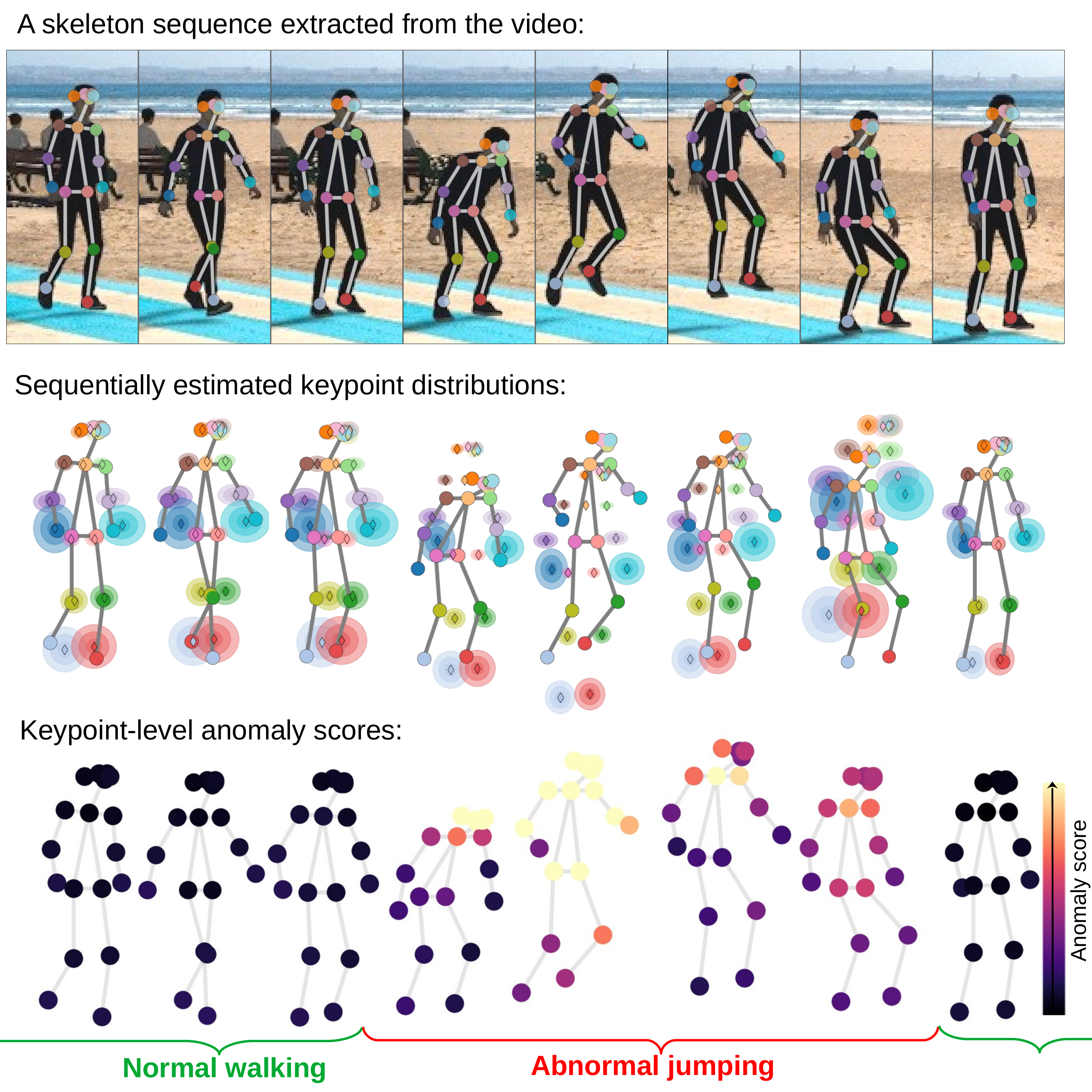}
    \caption{Qualitative performance of SeeKer on an example from UBnormal. During normal behavior, the observed keypoints emerge at anticipated locations.
    Abnormal behavior causes significant deviation from the expected keypoint locations, which enables seamless anomaly detection.
    }
    \label{fig:normal_abnormal}
\end{figure}

\begin{figure}[htb]
\centering
\includegraphics[width=\linewidth]{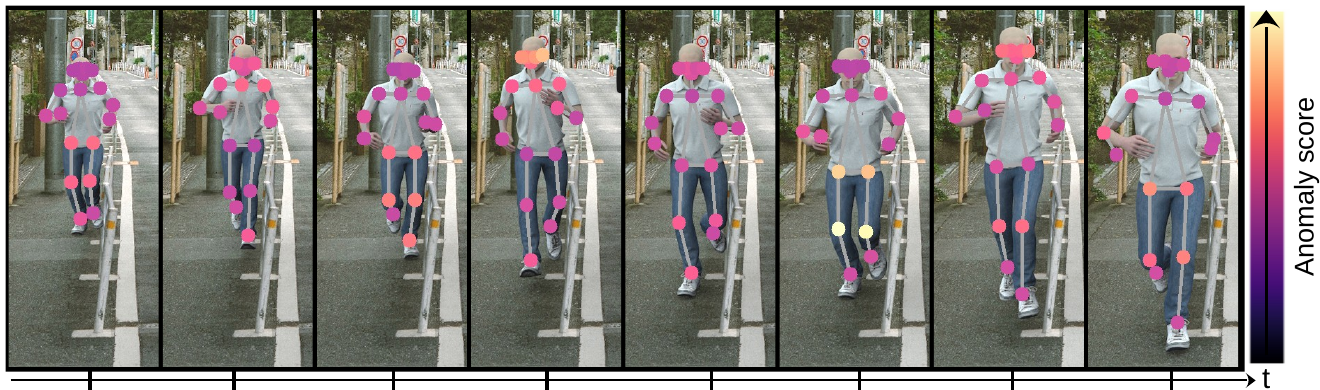}
\caption{
Interpretability of SeeKer  shown on an anomalous action from UBnormal (running).
SeeKer anomaly score aggregates per-keypoint likelihoods denoted with
varying color intensities.
Each keypoint contribution to the final score changes over time.}
\label{fig:explainability}
\end{figure}

The SeeKer anomaly score (\ref{eq:anomaly_score}) is highly interpretable, as it consolidates keypoint likelihoods into skeleton-level anomaly scores while operating in the original 2D space.
Figure~\ref{fig:normal_abnormal} shows 
the difference in predictions during normal and abnormal behavior.
The observations align closely with predictions of possible keypoint locations at normal timesteps.
At abnormal timesteps, the predictions tend to follow normal motion patterns and deviate significantly from actual observations, resulting in high anomaly scores.
In addition, Figure~\ref{fig:explainability} emphasizes the interpretability of per-keypoint contributions.
At each timestep, keypoints contribute with different intensity to the abnormality of the skeleton, as visualized through color variation.
Notably, this level of analysis is not feasible with previous  methods.

\begin{figure}[ht]
    \centering
    \includegraphics[width=\linewidth]{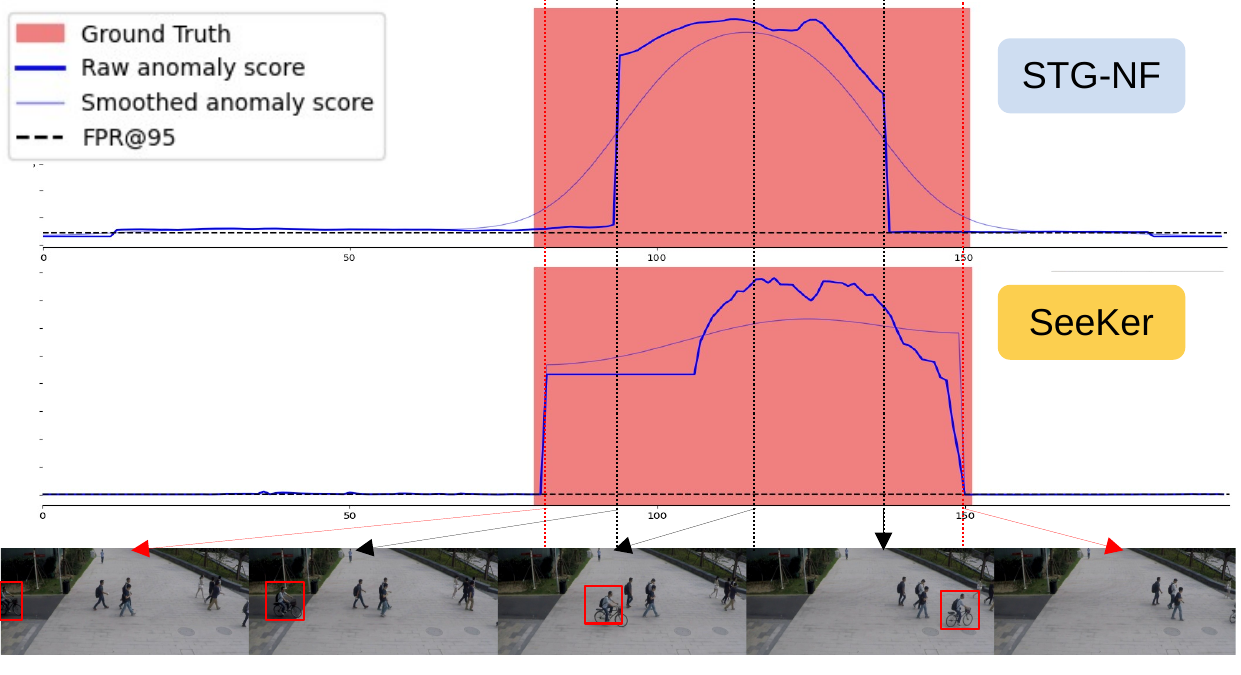}
    \caption{
    SeeKer accurately signals anomalous frames at the temporal border of the abnormal action, as shown by the blue curves, even without Gaussian smoothing (example from ShanghaiTech).}
    \label{fig:qual_comp}
\end{figure}

Figure \ref{fig:qual_comp}
compares SeeKer and the best skeleton-oriented baseline STG-NF \cite{hirschorn2023iccv} on an anomalous sequence from  ShanghaiTech (bike riding in the pedestrian area).
We plot both the raw anomaly scores and the anomaly scores smoothed with a 1D Gaussian filter.
The x- and y-axes represent time and anomaly score, respectively, and red-shaded regions mark frames labeled as anomalous.
SeeKer accurately identifies the anomalous frames, while STG-NF fails to detect anomalies at the temporal borders of the abnormal action.
Prior work  \cite{micorek24cvpr}
notes that ShanghaiTech contains noisy labels for skeleton-based anomaly detection.
For instance, a bicycle wheel may appear before the rider, delaying detection by skeleton-based methods.

\subsection{Discussion}
For all experiments in the discussion, we report  AUROC on the UBnormal and ShanghaiTech test subsetss.

\noindent
\textbf{On predefined or learned covariance.}
SeeKer predicts parameters of multivariate normal distribution, $\boldsymbol{\mu}$ and $\Sigma$.
We validate the importance of the learned covariance matrix compared to the predefined alternative.
We consider predefined covariance (identity matrix), learned diagonal covariance ($\Sigma_\theta = \boldsymbol{\sigma}^2 I$), and learned full covariance.
In the latter case, we model the inverse covariance matrix with Cholesky decomposition $\Sigma_\theta^{-1} =  L_\theta L_\theta^\top$ where $L_\theta$ is a lower triangular matrix outputted by the model \cite{dorta18cvpr}.
The decomposition ensures numerically stable full covariance implementation.
Note that the case with the identity matrix as covariance makes the second term of objective (\ref{eq:loss_objective}) constant.

Table \ref{tab:learned_cov} compares the performance of learned covariance, either full or diagonal, and baseline with the fixed covariance matrix. 
The obtained results indicate over 10\% absolute improvement with respect to the fixed covariance, emphasising the importance of the learned covariance in the SeeKer framework.
\begin{table}[ht]
\setlength{\tabcolsep}{12pt}
\footnotesize
\centering
  \begin{tabular}{c|cccc}
    \multirow{2}{*}{Covariance $\Sigma$} & \multicolumn{2}{c}{UBnormal} &  \multicolumn{2}{c}{ShanghaiTech}  \\
     & Full & HR & Full & HR  \\
    \hline
    identity & 61.4 & 63.4 & 74.1 &  74.4\\
    full & 77.8 & 78.1 & 84.3 & 85.8  \\
    diagonal & \textbf{77.9} & \textbf{78.9} &  \textbf{85.5} & \textbf{86.9} \\

\end{tabular}
    \caption{Learned covariance significantly outperforms the predefined identity matrix covariance.}
    \label{tab:learned_cov}
\end{table}

\noindent
\textbf{On prediction granularity.}
Autoregressive factorization can be spanned at different granularity levels, \textit{i.e.}~at the keypoint or skeleton level.
Each of these options requires the adjustment in masking strategy, as detailed in the Appendix \ref{app:architectures}.
Table \ref{tab:data_level} validates the performance of each of these granularities with
the same hyperparameters. 
We observe that modeling skeleton sequences at the keypoint level consistently outperforms its skeleton-level counterpart across the datasets.
Interestingly, 
both granularities from Table \ref{tab:data_level} outperform 
the current state of the art on UBnormal.
This outcome underscores the effectiveness of our autoregressive modelling of skeleton sequences.
\begin{table}[ht]
\setlength{\tabcolsep}{8pt}
\footnotesize
\centering
  \begin{tabular}{l|cccc}
    \multirow{2}{*}{Granularity} & \multicolumn{2}{c}{UBnormal} &  \multicolumn{2}{c}{ShanghaiTech}  \\
    & Full & HR & Full & HR  \\
    \hline
    \textbf{Keypoint} & \textbf{77.9} & \textbf{78.9} &  \textbf{85.5} & \textbf{86.9} \\
    Skeleton & 75.0 & 76.8 & 79.0 & 82.1\\
\end{tabular}
    \caption{Per-keypoint vs.~per-skeleton prediction granularity.}
    \label{tab:data_level}
\end{table}

\noindent
\textbf{On the impact of keypoint uncertainty.}
Our anomaly score weights conditional log-likelihoods of each keypoint
with its keypoint extractor uncertainty from the skeleton detector.
Table \ref{tab:confidence} shows that incorporating the keypoint uncertainty into the final anomaly score substantially enhances anomaly detection performance across all datasets.
These results emphasise that aggregating uncertainties of all pipeline components contribute to the end decision.  
Note that previous best methods \cite{hirschorn2023iccv, micorek24cvpr} cannot easily incorporate keypoint extraction uncertainty in their anomaly score. 
We also integrate the detector uncertainty in the training process as discussed in Appendix \ref{app:discusion}.
\begin{table}[ht]
\setlength{\tabcolsep}{5pt}
\footnotesize
\centering
  \begin{tabular}{cc|cccc}
    Confidence & Anomaly & \multicolumn{2}{c}{UBnormal} & \multicolumn{2}{c}{ShanghaiTech}  \\
    weighting  & score  & Full & HR & Full & HR  \\
    \hline
    \ding{55} & Eq.~(\ref{eq:skeleton_score_inial}) & 76.8 & 77.7 & 83.7 & 85.2\\
    \checkmark & Eq.~(\ref{eq:anomaly_score}) & \textbf{77.9} & \textbf{78.9} &  \textbf{85.5} & \textbf{86.9}
\end{tabular}
    \caption{Integrating keypoint detection uncertainty into the anomaly score improves performance.}
    \label{tab:confidence}
\end{table}

\noindent
\textbf{On the heuristic smoothing.}
Post-process smoothing is a common technique for reducing outliers in time-sequence data. Previous works \cite{hirschorn2023iccv, flaborea2023iccv, micorek24cvpr} apply Gaussian kernel smoothing in a sliding window to refine per-frame anomaly scores. 
This postprocessing strategy yields around three percentage points performance improvement for the STG-NF baseline.
SeeKer, on the other hand, accurately detects action boundaries by construction, 
as shown in Figure \ref{fig:qual_comp}.
Thus, smoothing contributes less to the end score.
Furthermore, if we remove postprocessing from both methods, SeeKer consistently outperforms STG-NF on all datasets. 
\begin{table}[ht]
\setlength{\tabcolsep}{5pt}
\footnotesize
\centering
  \begin{tabular}{cc|cccc}
    Method & Smoothing & \multicolumn{2}{c}{UBnormal} & \multicolumn{2}{c}{ShanghaiTech}  \\
      &   & Full & HR & Full & HR  \\
    \hline
    STG-NF & \ding{55} & 68.2 & 68.0 & 83.0 & 84.4  \\
    SeeKer & \ding{55} & \textbf{76.8} & \textbf{77.7} & \textbf{83.4} &  \textbf{84.9}\\[0.5em]
    STG-NF & \checkmark & 71.8 & 73.9 & \textbf{85.9} & \textbf{87.4} \\
    SeeKer & \checkmark & \textbf{77.9} & \textbf{78.9} &  85.5 & 86.9
\end{tabular}
    \caption{Post-process smoothing only slightly boosts the performance of SeeKer. 
    Importantly, SeeKer without smoothing outperforms STG-NF with smoothing. }
    \label{tab:smoothing}
\end{table}

\noindent
\textbf{On the keypoint ordering.}
SeeKer assumes a fixed ordering of skeleton keypoints. 
Ablations with different keypoint orderings incur a variance of less than 0.1, demonstrating SeeKer invariance to predefined keypoint order. Thus, we rely on the standard keypoint ordering \cite{lin14eccv} by default.

\noindent
\textbf{On autoregressive model architectures.}
We analyse the performance of SeeKer depending on the specific model architecture used to predict parameters $\boldsymbol\mu$ and $\Sigma$.
In addition to the causally masked fully-connected model, we also test transformer architecture \cite{vaswani17nips} that use causal self-attention layers.
Each transformer token is a 2D skeleton keypoint location projected into a higher dimensional space.
Further, we encode timestep and keypoint IDs to positional encoding of each token, as detailed in Appendix \ref{app:architectures}.

Table \ref{tab:attention} shows that the masked fully connected model significantly outperforms its transformer-based counterpart.
We attribute this gap to the low variability in skeleton sequences, as each person's gait differs only subtly.
Additionally, skeleton sequences are much shorter than text paragraphs, and keypoint tokens have have a much lower dimensionality, which influences the nature of the task and applicability of simpler models.
\begin{table}[ht]
\setlength{\tabcolsep}{8pt}
\footnotesize
\centering
  \begin{tabular}{l|cccc}
    \multirow{2}{*}{Architecture} & \multicolumn{2}{c}{UBnormal} &  \multicolumn{2}{c}{ShanghaiTech} \\
    & Full & HR & Full & HR  \\
    \hline
    Transformer decoder \cite{vaswani17nips} & 71.1 & 71.6 & 79.1 & 80.0 \\
 Masked fully connected \cite{germain15icml} & \textbf{77.9} & \textbf{78.9} &  \textbf{85.5} & \textbf{86.9} \\

\end{tabular}
    \caption{Comparison of the proposed masked fully-connected model  and the causal transformer decoder.}
    \label{tab:attention}
\end{table}


%% file: sec/5_conclusion.tex
\section{Conclusion}
\label{sec:conclusion}


We introduced SeeKer, a detector for anomalous human behaviour based on generative modeling of skeletal sequences.
Unlike previous approaches of this kind,
SeeKer considers each skeleton as a collection of keypoints and models sequence density as a product of Gaussian conditional keypoint densities.
The parameters of the conditionals
$(\boldsymbol\mu, \Sigma)$
are regressed by a causal deep model that ensures that each prediction can only observe the locations of the precedent keypoints.
The proposed architecture infers all per-keypoint distributions
with a single forward pass.
We implement the corresponding causal architecture with masked fully connected layers.
Interestingly, this formulation outperforms its counterpart based on a transformer decoder,
probably due to the specifics of our data.
During inference, we evaluate the per-keypoint distributions of the considered skeleton given the preceding keypoints in a sliding window, and summarize them into a skeleton-level anomaly score.
Here, we take advantage of the composite structure of the proposed anomaly score by weighting the per-keypoint log-densities with the corresponding confidence values of the keypoint detector.
The resulting anomaly score
identifies abnormal behaviour
by assessing the surprise
of our model.
Furthermore, our compositional anomaly score is highly
interpretable since we can track keypoint contributions to the final decision.
The conducted experiments indicate
that SeeKer surpasses the state of the art
on UBnormal and MSAD-HR by a wide margin,
while delivering a close
runner-up performance on ShanghaiTech.





%% file: sec/6_acknowledgement.tex
\section{Acknowledgements}
\label{sec:acknowledgements}
This work has been supported by Croatian Recovery and Resilience Fund -
NextGenerationEU (grant C1.4 R5-I2.01.0001). The authors declare no competing interests.

%% file: appendix_content.tex
\appendix
\newpage
\onecolumn

\begin{center}
  {\Large \textbf{Sequential keypoint density estimator: \\
an overlooked baseline of skeleton-based video anomaly detection}} \\[1em]
    {\large Supplementary Material}
\end{center}

\section{Derivation of the SeeKer optimization objective}
\label{app:derivation}

We train the SeeKer model by minimizing the sequence likelihood (\ref{eq:loss_objective}) that boils down to minimizing the Mahalanobis distance between the observed keypoint $X_{t, n}$ and the multivariate normal distribution with parameters $\boldsymbol\mu_\theta$ and $\Sigma_\theta$, alongside a regularization term.
The step-by-step derivation goes as follows: 
\begin{align}
    L(\theta; \mathcal{D}) &= -\sum_{\mathbf{X} \in \mathcal{D}} \ln p_\theta(\mathbf{X}) \\
    &= -\sum_{\mathbf{X} \in \mathcal{D}} \sum_{t=1}^T  \sum_{n=1}^N \ln \mathcal{N}(X_{t,n}|\boldsymbol{\mu}_\theta, \Sigma_\theta) \\
    &= - \sum_{\mathbf{X}, t, n} \ln  \frac{1}{(2\pi)  |\Sigma_\theta|^{1/2}} \exp \left[ -\frac{1}{2} (\mathbf{X}_{t,n} - \boldsymbol{\mu}_\theta)^T \Sigma_\theta^{-1} (\mathbf{X}_{t,n} - \boldsymbol{\mu}_\theta) \right] \\
    &= \frac{1}{2} \sum_{\mathbf{X}, t, n} 
    \left[ (\mathbf{X}_{t,n} - \boldsymbol{\mu}_\theta)^T \Sigma_\theta^{-1} (\mathbf{X}_{t,n} - \boldsymbol{\mu}_\theta) \right] + \ln (2\pi) + \ln \det \Sigma_\theta \\
    &\cong \sum_{\mathbf{X}, t, n} 
    \left[ (\mathbf{X}_{t,n} - \boldsymbol{\mu}_\theta)^T \Sigma_\theta^{-1} (\mathbf{X}_{t,n} - \boldsymbol{\mu}_\theta) \right] + \ln \det \Sigma_\theta 
\end{align}
In practice, our main model used diagonal covariance, $\Sigma=\boldsymbol\sigma I$ which yields:
\begin{equation}
    L(\theta; \mathcal{D}) = \sum_{\mathbf{X}, t, n} 
    \left[ (\mathbf{X}_{t,n} - \boldsymbol{\mu}_\theta)^T \text{diag}(1/\boldsymbol\sigma) (\mathbf{X}_{t,n} - \boldsymbol{\mu}_\theta) \right] + \ln \sigma_x + \ln \sigma_y
\end{equation}

\section{Architectures of autoreregressive density estimators}
\label{app:architectures}

\subsection{Causal fully connected model}
Causality constraints can be effectively incorporated into deep fully connected models by strategically masking weight matrices \cite{germain15icml,bengio99neurips}. 
Given an input vector $\mathbf{x}$ of size $T \times N \times D$ (e.g.~a flattened matrix $\mathbf{X}$ containing a skeleton sequence), we define the causal fully connected layer (C-FC$_l$) as:
\begin{equation}
    \text{C-FC$_l$}(\mathbf{x}) = (\mathbf{W} \odot \mathbf{M}) \cdot \mathbf{x} + \mathbf{b} .
\end{equation}
The mask $\mathbf{M}$ enforces causality by setting weights that correspond to current and subsequent input dimensions to zero.
By stacking multiple C-FC$_l$  layers, we can construct a deep causal fully connected model (C-FC).
Assuming $L$ hidden layers, we can express the C-FC with the following equations:
\begin{align}
\mathbf{h}_{0} &= \mathbf{x} \\
    \mathbf{h}_{l} &= g((\mathbf{W}_l \odot \mathbf{M}_l)\mathbf{h}_{l-1} + \mathbf{b}_l), \,\, l=\{1, \dots L\}\\
    \boldsymbol\mu,  \boldsymbol\sigma &= (\mathbf{W}_o \odot \mathbf{M}_o)\mathbf{h}_{L} + \mathbf{b}_o
\end{align}
Here, $g$ is some non-linear activation function, such as ReLU.
We mask the weights of intermediate layers $l$ with blockwise lower triangular matrices $\mathbf{M}^l$: 
\begin{equation}
    \mathbf{M}^{l}_{i,j} =
    \begin{cases} 
        1, & \text{if } \lfloor j / D \rfloor \leq \lfloor i / D \rfloor \\ 
        0, & \text{otherwise}
    \end{cases}
    \label{eq:M_in}
\end{equation}
ensuring that the elements of each keypoint are still codependent.
The mask of the output layer $\mathbf{M}^o$ is an upper diagonal 
\begin{equation}
    \mathbf{M}^o_{i,j} =
    \begin{cases} 
        1, & \text{if } \lfloor j / D \rfloor > \lfloor i / D \rfloor \\ 
        0, & \text{otherwise},
    \end{cases}
    \label{eq:M_out}
\end{equation}
to ensure independent predictions, e.g.~ computation without direct influence from its own current and all succeeding representation.
We duplicate the mask in the output layer since we output two values for each input, e.g.~ $\boldsymbol\mu$ and $\Sigma$.
For the main experiments that are on the keypoint granularity level, we use $D=2$, and
for the experiments on the skeleton granularity level $D=N \cdot T = 36$ (Table~\ref{tab:data_level}).
Finally, to ensure that all keypoints contribute equally, the hidden layer dimensions must be a multiple of the input dimension. This guarantees a balanced flow of information across all keypoints during processing.

\begin{figure}[htb]
    \centering
    \includegraphics[width=0.5\linewidth]{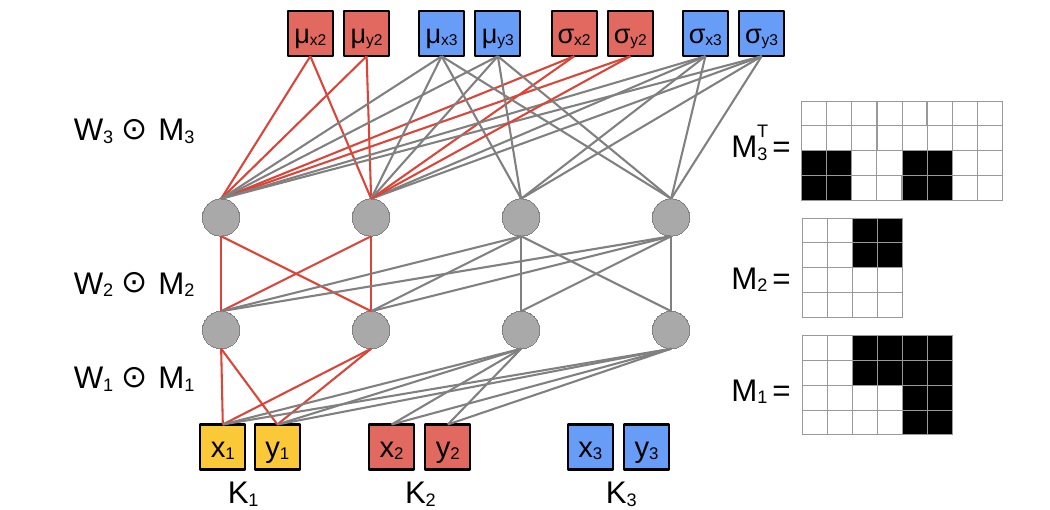}
    \caption{A minimal example of the causal deep fully connected model. 
    Here we causally predict parameters of the distribution for target keypoint K$_2$ with respect to the preceding keypoint K$_1$ and another target keypoint K$_3$ with respect to preceding K$_1$ and K$_2$.
    We also show examples of masks that ensure causality.}
    \label{fig:c-mlp}
\end{figure}

Figure \ref{fig:c-mlp} shows a minimal example of a causal fully-connected model.
For example, the predicted parameters of distribution under which the keypoint K$_2$ (red) should be probable depends on the keypoint K$_1$ (blue) as highlighted with red weights (see Figure 3 in the main text). 
To maintain clarity, keypoints of preceding skeletons and bias terms $\mathbf{b}_l$ are omitted from the visualization.


\subsection{Causal transformer}
Transformer decoder with causal self-attention layers \cite{vaswani17nips, brown20neurips} is an alternative architecture that admits autoregressive factorization by construction.
We briefly review the causal self-attention layer and analyse its shortcomings in the context of skeleton sequences.

The causal self-attention layer (C-SA) processes each token in a sequence $\mathbf{X}$ by attending only to the preceding tokens.
In the case of skeleton keypoint sequences, the corresponding input is an $L \times d$ matrix that collects $L$ $d$-dimensional tokens.
This transformation
is parametrized by three learnable projection matrices, $\mathbf{W}_q, \mathbf{W}_k$ and $\mathbf{W}_v$ 
which generate the query, key, and value representations, respectively, as follows:
\begin{equation}
    \text{C-SA}(\mathbf{X}) =  \left(\sigma (\mathbf{X} \mathbf{W}_q \mathbf{W}_k^\top \mathbf{X}^\top / \sqrt{d}) \odot \mathbf{M}^{\text{SA}}\right) \mathbf{X} \mathbf{W}_v .
\end{equation}
Here, $\sigma$ represents the row-wise softmax function, while the 
mask $M$ ensures causality by not attending the future tokens in the sequence: 
\begin{equation}
    \mathbf{M}^{\text{SA}}_{ij} =
    \begin{cases} 
        1, & \text{if } j \leq i \\ 
        0, & \text{otherwise}.
    \end{cases}
\end{equation}
This design ensures all model parameters are involved in predicting next sequence token.

Our empirical observations indicate that the transformer architecture is a suboptimal design choice for autoregressive density estimation on skeleton sequences.

\section{Additional discussion}
\label{app:discusion}

\subsection{On hyperparameter sensitivity.}
We tune the SeeKer hyperparameters on the validation subset of UBnormal \cite{acsintoae22cvpr}, and apply the same settings to ShanghaiTech and MSAD.
We use the early stopping criteria and optimize for maximum validation AUROC.
Table \ref{tab:validation} reports the performance of SeeKer across key hyperparameters: skeleton sequence length, number of hidden layers, and the input expansion factor that governs the size of the hidden layers.
SeeKer consistently achieves competitive performance on the UBnormal validation set across different hyperparameter choices.
\begin{table}[ht]
\setlength{\tabcolsep}{2pt}
\footnotesize
\centering
  \begin{tabular}{cc|cc|cc}
    \makecell{Sequence \\ length} & AUROC & \makecell{Nr. hidden \\ layers} & AUROC & \makecell{Expansion \\ factor} & AUROC  \\
    \hline
    8 & 84.9 & 1 & 84.8 & 1 & 85.4 \\
    16 & 84.1 & 2 & 84.8 & 2 & 85.4 \\
    \textbf{24} & \textbf{85.6} & \textbf{3} & \textbf{85.6} & 3 & 85.6 \\
    32 & 85.5 & 4 & 84.6 & \textbf{4} & \textbf{85.6}\\
    48 & 85.5 & 5 & 80.1 & 5 & 85.5 \\
\end{tabular}
    \caption{Validation of the model architecture hyperparameters on the UBnormal validation set shows that minor adjustments in hyperparameter selection have minimal impact on performance.}
    \label{tab:validation}
\end{table}

\subsection{On integrating keypoint detection confidence.}
Additionally, during training, we train only on confident skeletons. 
We leverage the per-keypoint detection confidence and define the skeleton confidence as the mean confidence over its keypoints.
We filter out skeletons with a detection confidence less than 0.4. 
This includes occluded skeletons, or skeletons at the border of the frame. Table~\ref{tab:training_confidence} validates the importance on training only on skeletons with high detection confidence.
\begin{table}[ht]
\setlength{\tabcolsep}{5pt}
\footnotesize
\centering
  \begin{tabular}{cc|cccc}
    Confidence & confidence  & \multicolumn{2}{c}{UBnormal} & \multicolumn{2}{c}{ShangahaiTech}  \\
    weighting  & filtering  & Full & HR & Full & HR  \\
    \hline
    \ding{55} & Eq.~(\ref{eq:skeleton_score_inial}) & 75.5 & 76. & 83.7 & 84.3\\
    \checkmark & Eq.~(\ref{eq:anomaly_score}) & \textbf{77.9} & \textbf{78.9} &  \textbf{85.5} & \textbf{86.9}
\end{tabular}
    \caption{Training on skeletons with high detection confidence improves the performance in terms of AUROC on UBnormal and ShanghaiTech test.}
    \label{tab:training_confidence}
\end{table}

\subsection{On non-human related anomalies.}
\begin{figure*}[htb]
    \centering
    \includegraphics[width=0.7\linewidth]{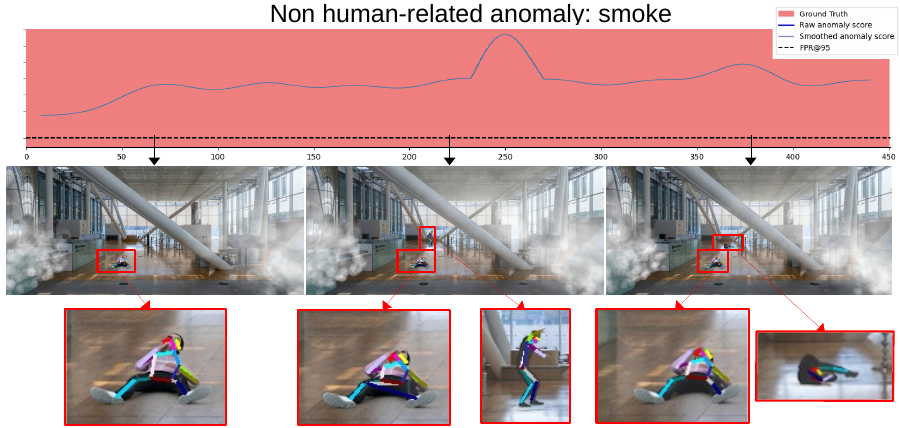}
    \caption{SeeKer can signal non-human related anomalies (e.g. smoke) as long as there are humans in the scene, since humans strike unusual poses in such cases (example from UBnormal).}
    \label{fig:non_human}
\end{figure*}

Figure \ref{fig:non_human} illustrates SeeKer anomaly scores for an abnormal event from the UBnormal dataset labeled as non-human related (smoke).
SeeKer accurately flags the corresponding anomalous frames since people exhibit unusual poses in response to this anomaly.
Some works on skeleton-based methods \cite{flaborea2023iccv, rodrigues2020wacv} tend to remove non-human related anomalies from the test dataset. 
However, these test cases highlight the versatility of skeleton-based methods like SeeKer by demonstrating their ability to detect anomalies even when they are only indirectly related to human behavior.
Moreover, such testing scenarios regularly appear in relevant benchmarks \cite{acsintoae22cvpr,liu18cvpr,zhu25neurips}.

\subsection{On computational requirements.}
Our method is efficient in terms of computational requirements. Training requires only 0.76 GB of GPU memory and completes in under 10 minutes on a single NVIDIA GeForce RTX 3090. 

\section{Limitations}
\label{app:limitations}

\noindent
\textbf{Keypoint extractors.}~Contemporary skeleton extractors \cite{fang23tpami} and trackers \cite{xiu18bmvc} can struggle in challenging scenarios, such as dense crowds \cite{lu13iccv} and poor lighting conditions \cite{sultani18cvpr}.
This may potentially limit the effectiveness of SeeKer in these cases. 
However, future advances in skeleton extractions can be easily incorporated into our framework and thus enhance SeeKer applicability across diverse conditions.

\noindent
\textbf{Potential biases.}~SeeKer relies on skeleton sequences that are invariant to human-related appearance. 
Consequently, we cannot introduce any appearance-related biases but may inherit biases of skeleton extractors.

%% file: main.bbl
\begin{thebibliography}{62}
\providecommand{\natexlab}[1]{#1}
\providecommand{\url}[1]{\texttt{#1}}
\expandafter\ifx\csname urlstyle\endcsname\relax
  \providecommand{\doi}[1]{doi: #1}\else
  \providecommand{\doi}{doi: \begingroup \urlstyle{rm}\Url}\fi

\bibitem[md1(2018)]{md18ssa}
Reprint of: Mahalanobis, p.c. (1936) ``on the generalised distance in statistics.''.
\newblock \emph{Sankhya Ser. A}, 2018.

\bibitem[Abati et~al.(2019)Abati, Porrello, Calderara, and Cucchiara]{abati2019cvpr}
Davide Abati, Angelo Porrello, Simone Calderara, and Rita Cucchiara.
\newblock Latent space autoregression for novelty detection.
\newblock In \emph{{IEEE} Conference on Computer Vision and Pattern Recognition, {CVPR} 2019, Long Beach, CA, USA, June 16-20, 2019}, 2019.

\bibitem[Acsintoae et~al.(2022)Acsintoae, Florescu, Georgescu, Mare, Sumedrea, Ionescu, Khan, and Shah]{acsintoae22cvpr}
Andra Acsintoae, Andrei Florescu, Mariana{-}Iuliana Georgescu, Tudor Mare, Paul Sumedrea, Radu~Tudor Ionescu, Fahad~Shahbaz Khan, and Mubarak Shah.
\newblock Ubnormal: New benchmark for supervised open-set video anomaly detection.
\newblock In \emph{Computer Vision and Pattern Recognition}, 2022.

\bibitem[Bengio and Bengio(1999)]{bengio99neurips}
Yoshua Bengio and Samy Bengio.
\newblock Modeling high-dimensional discrete data with multi-layer neural networks.
\newblock In \emph{Advances in Neural Information Processing Systems}. MIT Press, 1999.

\bibitem[Berezina et~al.(2010)Berezina, Rudoy, and Wolfe]{berezina10icassp}
Maria~A. Berezina, Daniel Rudoy, and Patrick~J. Wolfe.
\newblock Autoregressive modeling of voiced speech.
\newblock In \emph{{IEEE} International Conference on Acoustics, Speech, and Signal Processing, {ICASSP}}, pages 5042--5045, 2010.

\bibitem[Braverman(2015)]{braverman15book}
Vladimir Braverman.
\newblock \emph{Sliding Window Algorithms}.
\newblock 2015.

\bibitem[Brown et~al.(2020)Brown, Mann, Ryder, Subbiah, Kaplan, Dhariwal, Neelakantan, Shyam, Sastry, Askell, Agarwal, Herbert{-}Voss, Krueger, Henighan, Child, Ramesh, Ziegler, Wu, Winter, Hesse, Chen, Sigler, Litwin, Gray, Chess, Clark, Berner, McCandlish, Radford, Sutskever, and Amodei]{brown20neurips}
Tom~B. Brown, Benjamin Mann, Nick Ryder, Melanie Subbiah, Jared Kaplan, Prafulla Dhariwal, Arvind Neelakantan, Pranav Shyam, Girish Sastry, Amanda Askell, Sandhini Agarwal, Ariel Herbert{-}Voss, Gretchen Krueger, Tom Henighan, Rewon Child, Aditya Ramesh, Daniel~M. Ziegler, Jeffrey Wu, Clemens Winter, Christopher Hesse, Mark Chen, Eric Sigler, Mateusz Litwin, Scott Gray, Benjamin Chess, Jack Clark, Christopher Berner, Sam McCandlish, Alec Radford, Ilya Sutskever, and Dario Amodei.
\newblock Language models are few-shot learners.
\newblock In \emph{Advances in Neural Information Processing Systems 33: Annual Conference on Neural Information Processing Systems 2020, NeurIPS 2020, December 6-12, 2020, virtual}, 2020.

\bibitem[Bărbălău et~al.(2022)Bărbălău, Ionescu, Georgescu, Dueholm, Ramachandra, Nasrollahi, Khan, Moeslund, and Shah]{barbalau2022cviu}
Antonio Bărbălău, Radu~Tudor Ionescu, Mariana-Iuliana Georgescu, Jacob~Velling Dueholm, Bharathkumar Ramachandra, Kamal Nasrollahi, Fahad~Shahbaz Khan, Thomas~Baltzer Moeslund, and Mubarak Shah.
\newblock Ssmtl++: Revisiting self-supervised multi-task learning for video anomaly detection.
\newblock \emph{Comput. Vis. Image Underst.}, page 103656, 2022.

\bibitem[Carion et~al.(2020)Carion, Massa, Synnaeve, Usunier, Kirillov, and Zagoruyko]{carion20eccv}
Nicolas Carion, Francisco Massa, Gabriel Synnaeve, Nicolas Usunier, Alexander Kirillov, and Sergey Zagoruyko.
\newblock End-to-end object detection with transformers.
\newblock In \emph{European Conference on Computer Vision}, 2020.

\bibitem[Chen et~al.(2020)Chen, Tian, and He]{chen20cviu}
Yucheng Chen, Yingli Tian, and Mingyi He.
\newblock Monocular human pose estimation: {A} survey of deep learning-based methods.
\newblock \emph{Comput. Vis. Image Underst.}, 192:\penalty0 102897, 2020.

\bibitem[de~Morais et~al.(2019)de~Morais, Le, Tran, Saha, Mansour, and Venkatesh]{morais2019cvpr}
Romero F. A.~B. de Morais, Vuong Le, Truyen Tran, Budhaditya Saha, Moussa~Reda Mansour, and Svetha Venkatesh.
\newblock Learning regularity in skeleton trajectories for anomaly detection in videos.
\newblock In \emph{{IEEE} Conference on Computer Vision and Pattern Recognition, {CVPR} 2019, Long Beach, CA, USA, June 16-20, 2019}, 2019.

\bibitem[Devlin et~al.(2019)Devlin, Chang, Lee, and Toutanova]{devlin19naacl}
Jacob Devlin, Ming{-}Wei Chang, Kenton Lee, and Kristina Toutanova.
\newblock {BERT:} pre-training of deep bidirectional transformers for language understanding.
\newblock In \emph{Conference of the North American Chapter of the Association for Computational Linguistics: Human Language Technologies, {NAACL-HLT}}, 2019.

\bibitem[Dorta et~al.(2018)Dorta, Vicente, Agapito, Campbell, and Simpson]{dorta18cvpr}
Garoe Dorta, Sara Vicente, Lourdes Agapito, Neill D.~F. Campbell, and Ivor Simpson.
\newblock Structured uncertainty prediction networks.
\newblock In \emph{2018 {IEEE} Conference on Computer Vision and Pattern Recognition, {CVPR} 2018, Salt Lake City, UT, USA, June 18-22, 2018}, 2018.

\bibitem[dos Santos et~al.(2019)dos Santos, Ribeiro, and Ponti]{santos19vcir}
Fernando~Pereira dos Santos, Leonardo Sampaio~Ferraz Ribeiro, and Moacir~A. Ponti.
\newblock Generalization of feature embeddings transferred from different video anomaly detection domains.
\newblock \emph{J. Vis. Commun. Image Represent.}, 2019.

\bibitem[Fang et~al.(2023)Fang, Li, Tang, Xu, Zhu, Xiu, Li, and Lu]{fang23tpami}
Haoshu Fang, Jiefeng Li, Hongyang Tang, Chao Xu, Haoyi Zhu, Yuliang Xiu, Yong{-}Lu Li, and Cewu Lu.
\newblock Alphapose: Whole-body regional multi-person pose estimation and tracking in real-time.
\newblock \emph{{IEEE} Trans. Pattern Anal. Mach. Intell.}, 2023.

\bibitem[Flaborea et~al.(2023)Flaborea, Collorone, di~Melendugno, D'Arrigo, Prenkaj, and Galasso]{flaborea2023iccv}
Alessandro Flaborea, Luca Collorone, Guido Maria~D'Amely di Melendugno, Stefano D'Arrigo, Bardh Prenkaj, and Fabio Galasso.
\newblock Multimodal motion conditioned diffusion model for skeleton-based video anomaly detection.
\newblock In \emph{{IEEE/CVF} International Conference on Computer Vision, {ICCV} 2023, Paris, France, October 1-6, 2023}, 2023.

\bibitem[Flaborea et~al.(2024)Flaborea, di~Melendugno, D'Arrigo, Sterpa, Sampieri, and Galasso]{flaborea2024pr}
Alessandro Flaborea, Guido Maria~D'Amely di Melendugno, Stefano D'Arrigo, Marco~Aurelio Sterpa, Alessio Sampieri, and Fabio Galasso.
\newblock Contracting skeletal kinematics for human-related video anomaly detection.
\newblock \emph{Pattern Recognit.}, 2024.

\bibitem[Ge et~al.(2021)Ge, Liu, Wang, Li, and Sun]{ge21arxiv}
Zheng Ge, Songtao Liu, Feng Wang, Zeming Li, and Jian Sun.
\newblock Yolox: Exceeding yolo series in 2021.
\newblock \emph{ArXiv}, 2021.

\bibitem[Georgescu et~al.(2022)Georgescu, Ionescu, Khan, Popescu, and Shah]{georgescu2022tpami}
Mariana{-}Iuliana Georgescu, Radu~Tudor Ionescu, Fahad~Shahbaz Khan, Marius Popescu, and Mubarak Shah.
\newblock A background-agnostic framework with adversarial training for abnormal event detection in video.
\newblock \emph{{IEEE} Trans. Pattern Anal. Mach. Intell.}, 2022.

\bibitem[Georgescu et~al.(2020)Georgescu, Bărbălău, Ionescu, Khan, Popescu, and Shah]{georgescu2020cvpr}
Mariana-Iuliana Georgescu, Antonio Bărbălău, Radu~Tudor Ionescu, Fahad~Shahbaz Khan, Marius~Claudiu Popescu, and Mubarak Shah.
\newblock Anomaly detection in video via self-supervised and multi-task learning.
\newblock \emph{2021 IEEE/CVF Conference on Computer Vision and Pattern Recognition (CVPR)}, 2020.

\bibitem[Germain et~al.(2015)Germain, Gregor, Murray, and Larochelle]{germain15icml}
Mathieu Germain, Karol Gregor, Iain Murray, and Hugo Larochelle.
\newblock {MADE:} masked autoencoder for distribution estimation.
\newblock In \emph{International Conference on Machine Learning}, 2015.

\bibitem[Hasan et~al.(2016)Hasan, Choi, Neumann, Roy{-}Chowdhury, and Davis]{hasan2016cvpr}
Mahmudul Hasan, Jonghyun Choi, Jan Neumann, Amit~K. Roy{-}Chowdhury, and Larry~S. Davis.
\newblock Learning temporal regularity in video sequences.
\newblock In \emph{2016 {IEEE} Conference on Computer Vision and Pattern Recognition, {CVPR} 2016, Las Vegas, NV, USA, June 27-30, 2016}, pages 733--742. {IEEE} Computer Society, 2016.

\bibitem[Hirschorn and Avidan(2023)]{hirschorn2023iccv}
Or Hirschorn and Shai Avidan.
\newblock Normalizing flows for human pose anomaly detection.
\newblock In \emph{International Conference on Computer Vision}, 2023.

\bibitem[Hochreiter and Schmidhuber(1997)]{hochreiter97nc}
Sepp Hochreiter and J{\"{u}}rgen Schmidhuber.
\newblock Long short-term memory.
\newblock \emph{Neural Comput.}, 1997.

\bibitem[Ionescu et~al.(2017)Ionescu, Smeureanu, Alexe, and Popescu]{ionescu2017iccv}
Radu~Tudor Ionescu, Sorina Smeureanu, Bogdan Alexe, and Marius Popescu.
\newblock Unmasking the abnormal events in video.
\newblock In \emph{{IEEE} International Conference on Computer Vision, {ICCV} 2017, Venice, Italy, October 22-29, 2017}, 2017.

\bibitem[Jain et~al.(2020)Jain, Sharma, Velmurugan, and Banerjee]{jain2020icpr}
Yashswi Jain, Ashvini~Kumar Sharma, Rajbabu Velmurugan, and Biplab Banerjee.
\newblock Posecvae: Anomalous human activity detection.
\newblock In \emph{25th International Conference on Pattern Recognition, {ICPR} 2020, Virtual Event / Milan, Italy, January 10-15, 2021}, 2020.

\bibitem[Kanu{-}Asiegbu et~al.(2022)Kanu{-}Asiegbu, Vasudevan, and Du]{miracle2022arxiv}
Asiegbu~Miracle Kanu{-}Asiegbu, Ram Vasudevan, and Xiaoxiao Du.
\newblock Bipoco: Bi-directional trajectory prediction with pose constraints for pedestrian anomaly detection.
\newblock \emph{CoRR}, abs/2207.02281, 2022.

\bibitem[Lamb et~al.(2016)Lamb, ALIAS PARTH~GOYAL, Zhang, Zhang, Courville, and Bengio]{lamb16neurips}
Alex~M Lamb, Anirudh~Goyal ALIAS PARTH~GOYAL, Ying Zhang, Saizheng Zhang, Aaron~C Courville, and Yoshua Bengio.
\newblock Professor forcing: A new algorithm for training recurrent networks.
\newblock In \emph{Advances in Neural Information Processing Systems}, 2016.

\bibitem[Lan et~al.(2023)Lan, Wu, Hu, and Hao]{lan23hms}
Gongjin Lan, Yu Wu, Fei Hu, and Qi Hao.
\newblock Vision-based human pose estimation via deep learning: {A} survey.
\newblock \emph{{IEEE} Trans. Hum. Mach. Syst.}, 2023.

\bibitem[Leyva et~al.(2017)Leyva, Sanchez, and Li]{leyva17tip}
Roberto Leyva, Victor Sanchez, and Chang{-}Tsun Li.
\newblock Video anomaly detection with compact feature sets for online performance.
\newblock \emph{{IEEE} Trans. Image Process.}, 2017.

\bibitem[Li et~al.(2021)Li, Chang, and Liu]{li2021neurocmp}
Nanjun Li, Faliang Chang, and Chunsheng Liu.
\newblock Human-related anomalous event detection via spatial-temporal graph convolutional autoencoder with embedded long short-term memory network.
\newblock \emph{Neurocomputing}, 2021.

\bibitem[Li et~al.(2024)Li, Huang, Ildiz, Rawat, and Oymak]{li24aistats}
Yingcong Li, Yixiao Huang, Muhammed~Emrullah Ildiz, Ankit~Singh Rawat, and Samet Oymak.
\newblock Mechanics of next token prediction with self-attention.
\newblock In \emph{International Conference on Artificial Intelligence and Statistics}, 2024.

\bibitem[Lin et~al.(2014)Lin, Maire, Belongie, Hays, Perona, Ramanan, Doll{\'a}r, and Zitnick]{lin14eccv}
Tsung-Yi Lin, Michael Maire, Serge Belongie, James Hays, Pietro Perona, Deva Ramanan, Piotr Doll{\'a}r, and C~Lawrence Zitnick.
\newblock Microsoft coco: Common objects in context.
\newblock In \emph{Computer vision--ECCV 2014: 13th European conference, zurich, Switzerland, September 6-12, 2014, proceedings, part v 13}, pages 740--755. Springer, 2014.

\bibitem[Liu et~al.(2018{\natexlab{a}})Liu, Luo, Lian, and Gao]{liu18cvpr}
Wen Liu, Weixin Luo, Dongze Lian, and Shenghua Gao.
\newblock Future frame prediction for anomaly detection - {A} new baseline.
\newblock In \emph{Conference on Computer Vision and Pattern Recognition}, 2018{\natexlab{a}}.

\bibitem[Liu et~al.(2018{\natexlab{b}})Liu, Luo, Lian, and Gao]{liu2018cvpr}
Wen Liu, Weixin Luo, Dongze Lian, and Shenghua Gao.
\newblock Future frame prediction for anomaly detection - {A} new baseline.
\newblock In \emph{2018 {IEEE} Conference on Computer Vision and Pattern Recognition, {CVPR} 2018, Salt Lake City, UT, USA, June 18-22, 2018}, pages 6536--6545. Computer Vision Foundation / {IEEE} Computer Society, 2018{\natexlab{b}}.

\bibitem[Liu et~al.(2021)Liu, Nie, Long, Zhang, and Li]{liu2021iccv}
Zhian Liu, Yongwei Nie, Chengjiang Long, Qing Zhang, and Guiqing Li.
\newblock A hybrid video anomaly detection framework via memory-augmented flow reconstruction and flow-guided frame prediction.
\newblock \emph{2021 IEEE/CVF International Conference on Computer Vision (ICCV)}, 2021.

\bibitem[Lu et~al.(2013)Lu, Shi, and Jia]{lu13iccv}
Cewu Lu, Jianping Shi, and Jiaya Jia.
\newblock Abnormal event detection at 150 {FPS} in {MATLAB}.
\newblock In \emph{{IEEE} International Conference on Computer Vision, {ICCV} 2013, Sydney, Australia, December 1-8, 2013}, 2013.

\bibitem[Luo et~al.(2017)Luo, Liu, and Gao]{luo2017iccv}
Weixin Luo, Wen Liu, and Shenghua Gao.
\newblock A revisit of sparse coding based anomaly detection in stacked {RNN} framework.
\newblock In \emph{International Conference on Computer Vision}, 2017.

\bibitem[Luo et~al.(2020)Luo, Liu, and Gao]{luo2020neurcomp}
Weixin Luo, Wen Liu, and Shenghua Gao.
\newblock Normal graph: Spatial temporal graph convolutional networks based prediction network for skeleton based video anomaly detection.
\newblock \emph{Neurocomputing}, 2020.

\bibitem[Malach(2024)]{malach24icml}
Eran Malach.
\newblock Auto-regressive next-token predictors are universal learners.
\newblock In \emph{International Conference on Machine Learning}, 2024.

\bibitem[Markovitz et~al.(2020)Markovitz, Sharir, Friedman, Zelnik{-}Manor, and Avidan]{markovitz2020cvpr}
Amir Markovitz, Gilad Sharir, Itamar Friedman, Lihi Zelnik{-}Manor, and Shai Avidan.
\newblock Graph embedded pose clustering for anomaly detection.
\newblock In \emph{2020 {IEEE/CVF} Conference on Computer Vision and Pattern Recognition, {CVPR} 2020, Seattle, WA, USA, June 13-19, 2020}, 2020.

\bibitem[Micorek et~al.(2024)Micorek, Possegger, Narnhofer, Bischof, and Kozinski]{micorek24cvpr}
Jakub Micorek, Horst Possegger, Dominik Narnhofer, Horst Bischof, and Mateusz Kozinski.
\newblock {MULDE:} multiscale log-density estimation via denoising score matching for video anomaly detection.
\newblock In \emph{{IEEE/CVF} Conference on Computer Vision and Pattern Recognition, {CVPR} 2024, Seattle, WA, USA, June 16-22, 2024}, 2024.

\bibitem[Mo et~al.(2014)Mo, Monga, and Bala]{mo14icip}
Xuan Mo, Vishal Monga, and Raja Bala.
\newblock Simultaneous sparsity model for multi-perspective video anomaly detection.
\newblock In \emph{International Conference on Image Processing}, 2014.

\bibitem[Ramachandra and Jones(2020)]{ramachandra20wacv}
Bharathkumar Ramachandra and Michael~J. Jones.
\newblock Street scene: {A} new dataset and evaluation protocol for video anomaly detection.
\newblock In \emph{{IEEE} Winter Conference on Applications of Computer Vision, {WACV} 2020, Snowmass Village, CO, USA, March 1-5, 2020}, 2020.

\bibitem[Rodrigues et~al.(2020)Rodrigues, Bhargava, Velmurugan, and Chaudhuri]{rodrigues2020wacv}
Royston Rodrigues, Neha Bhargava, Rajbabu Velmurugan, and Subhasis Chaudhuri.
\newblock Multi-timescale trajectory prediction for abnormal human activity detection.
\newblock In \emph{{IEEE} Winter Conference on Applications of Computer Vision, {WACV} 2020, Snowmass Village, CO, USA, March 1-5, 2020}, 2020.

\bibitem[Rumelhart et~al.(1986)Rumelhart, Hinton, and Williams]{rumelhart86n}
David~E. Rumelhart, Geoffrey~E. Hinton, and Ronald~J. Williams.
\newblock Learning representations by back-propagating errors.
\newblock \emph{Nature}, 1986.

\bibitem[Sapp et~al.(2010)Sapp, Toshev, and Taskar]{sapp10eccv}
Benjamin Sapp, Alexander Toshev, and Ben Taskar.
\newblock Cascaded models for articulated pose estimation.
\newblock In \emph{European Conference on Computer Vision}, 2010.

\bibitem[Sultani et~al.(2018)Sultani, Chen, and Shah]{sultani18cvpr}
Waqas Sultani, Chen Chen, and Mubarak Shah.
\newblock Real-world anomaly detection in surveillance videos.
\newblock In \emph{Conference on Computer Vision and Pattern Recognition}, 2018.

\bibitem[Sun et~al.(2012)Sun, Kohli, and Shotton]{sun12cvpr}
Min Sun, Pushmeet Kohli, and Jamie Shotton.
\newblock Conditional regression forests for human pose estimation.
\newblock In \emph{Computer Vision and Pattern Recognition}, 2012.

\bibitem[Tarzanagh et~al.(2023)Tarzanagh, Li, Zhang, and Oymak]{tarzanaghL23neurips}
Davoud~Ataee Tarzanagh, Yingcong Li, Xuechen Zhang, and Samet Oymak.
\newblock Max-margin token selection in attention mechanism.
\newblock In \emph{Neural Information Processing Systems}, 2023.

\bibitem[Vaswani et~al.(2017)Vaswani, Shazeer, Parmar, Uszkoreit, Jones, Gomez, Kaiser, and Polosukhin]{vaswani17nips}
Ashish Vaswani, Noam Shazeer, Niki Parmar, Jakob Uszkoreit, Llion Jones, Aidan~N. Gomez, Lukasz Kaiser, and Illia Polosukhin.
\newblock Attention is all you need.
\newblock In \emph{Neural Information Processing Systems}, 2017.

\bibitem[Waibel et~al.(1989)Waibel, Hanazawa, Hinton, Shikano, and Lang]{waibel89assp}
Alexander Waibel, Toshiyuki Hanazawa, Geoffrey~E. Hinton, Kiyohiro Shikano, and Kevin~J. Lang.
\newblock Phoneme recognition using time-delay neural networks.
\newblock \emph{{IEEE} Trans. Acoust. Speech Signal Process.}, 1989.

\bibitem[Wang et~al.(2022)Wang, Wang, Qin, Zhang, Bao, and Huang]{wang2022eccv}
Guodong Wang, Yunhong Wang, Jie Qin, Dongming Zhang, Xiuguo Bao, and Di Huang.
\newblock Video anomaly detection by solving decoupled spatio-temporal jigsaw puzzles.
\newblock In \emph{Computer Vision - {ECCV} 2022 - 17th European Conference, Tel Aviv, Israel, October 23-27, 2022, Proceedings, Part {X}}, 2022.

\bibitem[Xiu et~al.(2018)Xiu, Li, Wang, Fang, and Lu]{xiu18bmvc}
Yuliang Xiu, Jiefeng Li, Haoyu Wang, Yinghong Fang, and Cewu Lu.
\newblock Pose flow: Efficient online pose tracking.
\newblock In \emph{British Machine Vision Conference 2018, {BMVC} 2018, Newcastle, UK, September 3-6, 2018}, 2018.

\bibitem[Yan et~al.(2023)Yan, Zhang, Liu, Pang, and Wang]{yan23iccv}
Cheng Yan, Shiyu Zhang, Yang Liu, Guansong Pang, and Wenjun Wang.
\newblock Feature prediction diffusion model for video anomaly detection.
\newblock In \emph{{IEEE/CVF} International Conference on Computer Vision, {ICCV} 2023, Paris, France, October 1-6, 2023}, 2023.

\bibitem[Yan et~al.(2018)Yan, Xiong, and Lin]{yan18aaai}
Sijie Yan, Yuanjun Xiong, and Dahua Lin.
\newblock Spatial temporal graph convolutional networks for skeleton-based action recognition.
\newblock In \emph{Proceedings of the Thirty-Second {AAAI} Conference on Artificial Intelligence, (AAAI-18), the 30th innovative Applications of Artificial Intelligence (IAAI-18), and the 8th {AAAI} Symposium on Educational Advances in Artificial Intelligence (EAAI-18), New Orleans, Louisiana, USA, February 2-7, 2018}, 2018.

\bibitem[Yu et~al.(2018)Yu, Yin, and Zhu]{yu18ijcai}
Bing Yu, Haoteng Yin, and Zhanxing Zhu.
\newblock Spatio-temporal graph convolutional networks: {A} deep learning framework for traffic forecasting.
\newblock In \emph{Proceedings of the Twenty-Seventh International Joint Conference on Artificial Intelligence, {IJCAI} 2018, July 13-19, 2018, Stockholm, Sweden}, 2018.

\bibitem[Yu et~al.(2020)Yu, Wang, Cai, Zhu, Xu, Yin, and Kloft]{yu20acm}
Guang Yu, Siqi Wang, Zhiping Cai, En Zhu, Chuanfu Xu, Jianping Yin, and Marius Kloft.
\newblock Cloze test helps: Effective video anomaly detection via learning to complete video events.
\newblock In \emph{{MM} '20: The 28th {ACM} International Conference on Multimedia, Virtual Event / Seattle, WA, USA, October 12-16, 2020}. {ACM}, 2020.

\bibitem[Zaheer et~al.(2022)Zaheer, Mahmood, Khan, Seg{\`{u}}, Yu, and Lee]{zaheer2022cvpr}
Muhammad~Zaigham Zaheer, Arif Mahmood, Muhammad~Haris Khan, Mattia Seg{\`{u}}, Fisher Yu, and Seung{-}Ik Lee.
\newblock Generative cooperative learning for unsupervised video anomaly detection.
\newblock In \emph{{IEEE/CVF} Conference on Computer Vision and Pattern Recognition, {CVPR} 2022, New Orleans, LA, USA, June 18-24, 2022}, pages 14724--14734. {IEEE}, 2022.

\bibitem[Zhou et~al.(2019)Zhou, Yang, Cavallaro, and Xiang]{zhou19iccv}
Kaiyang Zhou, Yongxin Yang, Andrea Cavallaro, and Tao Xiang.
\newblock Omni-scale feature learning for person re-identification.
\newblock In \emph{2019 {IEEE/CVF} International Conference on Computer Vision, {ICCV} 2019, Seoul, Korea (South), October 27 - November 2, 2019}, 2019.

\bibitem[Zhu et~al.(2024)Zhu, Wang, Liu, Nie, Liu, and Li]{zhu24tpami}
Leyan Zhu, Zitian Wang, Si Liu, Xuecheng Nie, Luoqi Liu, and Bo Li.
\newblock Multi-person pose regression with distribution-aware single-stage models.
\newblock \emph{{IEEE} Trans. Pattern Anal. Mach. Intell.}, 2024.

\bibitem[Zhu et~al.(2025)Zhu, Wang, Raj, Gedeon, and Chen]{zhu25neurips}
Liyun Zhu, Lei Wang, Arjun Raj, Tom Gedeon, and Chen Chen.
\newblock Advancing video anomaly detection: A concise review and a new dataset.
\newblock \emph{Advances in Neural Information Processing Systems}, 37:\penalty0 89943--89977, 2025.

\end{thebibliography}
